\documentclass{article}

\usepackage[final,nonatbib]{nips_2017}

\usepackage[utf8]{inputenc} % allow utf-8 input
\usepackage[T1]{fontenc}    % use 8-bit T1 fonts
\usepackage{url}            % simple URL typesetting
\usepackage{booktabs}       % professional-quality tables
\usepackage{amsfonts}       % blackboard math symbols
\usepackage{nicefrac}       % compact symbols for 1/2, etc.
\usepackage{microtype}      % microtypography

\usepackage{graphicx}
\usepackage{amsmath,amssymb} % define this before the line numbering.
\usepackage{color}

\usepackage{stmaryrd}
\usepackage{mydefs}
\usepackage{algorithm}
\usepackage{algorithmic}
\usepackage{caption}
\usepackage{multirow}
\usepackage{verbatim}
\usepackage{subfig}
\usepackage{xcolor}
\usepackage{enumerate,paralist}
\usepackage{booktabs}
\usepackage{wrapfig}

\usepackage[title]{appendix}

\newtheorem{definition}{Definition}

\definecolor{orange}{rgb}{1,0.5,0}
\definecolor{bargreen}{RGB}{1,186,56}
\definecolor{barred}{RGB}{249,118,110}
\definecolor{barblue}{RGB}{97,157,253}

\definecolor{c1}{rgb}{0.3467,    0.5360,    0.6907}
\definecolor{c2}{rgb}{0.9153,    0.2816,    0.2878}
\definecolor{c3}{rgb}{0.4416,    0.7490,    0.4322}
\definecolor{c4}{rgb}{1.0000,    0.5984,    0.2000}

\def\GRUNAME{SP-GRU}

\usepackage[pagebackref=true,breaklinks=true,letterpaper=true,colorlinks,bookmarks=false]{hyperref}
\usepackage{array}
\usepackage{tabularx}

\graphicspath{{figs/}}

\title{Sampling-free Uncertainty Estimation in Gated Recurrent Units with Exponential Families}

\author{
	  \textbf{Seong Jae Hwang}\\
	  Department of Computer Sciences\\
	  University of Wisconsin - Madison\\
	  Madison, WI 53703 \\
	  \texttt{sjh@cs.wisc.edu}
	  \and
	  \textbf{Ronak Mehta}\\
	  Department of Computer Sciences\\
	  University of Wisconsin - Madison\\
	  Madison, WI 53703 \\
	  \texttt{ronakrm@cs.wisc.edu}
	  \AND
	  \textbf{Hyunwoo J. Kim}\\
	  Department of Computer Sciences\\
	  University of Wisconsin - Madison\\
	  Madison, WI 53703 \\
	  \texttt{hwkim@cs.wisc.edu} 
    \and
    \textbf{Vikas Singh}\\
    Department of Computer Sciences\\
        Department of Biostat. \& Med. Informatics\\
    University of Wisconsin - Madison\\
    Madison, WI 53703 \\
    \texttt{vsingh@biostat.wisc.edu} \\
}

\begin{document}

\maketitle

\begin{abstract}
%NEEDS TO BE 5 WORDS SHORTER
There has recently been a concerted effort to derive mechanisms in vision and machine learning systems to offer uncertainty estimates of the predictions they make. Clearly, there are enormous benefits to a system that is not only accurate but also has a sense for when it is not sure. Existing proposals center around Bayesian interpretations of modern deep architectures -- these are effective but can often be computationally demanding. We show how classical ideas in the literature on exponential families on probabilistic networks provide an excellent starting point to derive uncertainty estimates in Gated Recurrent Units (GRU). Our proposal directly quantifies uncertainty deterministically, without the need for costly sampling-based estimation. We demonstrate how our model can be used to quantitatively and qualitatively measure uncertainty in unsupervised image sequence prediction. To our knowledge, this is the first result describing sampling-free uncertainty estimation for powerful sequential models such as GRUs. 
\end{abstract}

\section{Introduction}
Recurrent Neural Networks (RNNs) have achieved state-of-the-art performance in various sequence prediction tasks
such as machine translation \cite{wu2016google,jozefowicz2016exploring,zaremba2014recurrent}, speech recognition \cite{hinton2015distilling,amodei2016deep},
language models \cite{cho2014learning,mikolov2010recurrent}, image and video captioning \cite{pan2016hierarchical,yu2016video,venugopalan2015sequence} as well as
medical applications \cite{jagannatha2016bidirectional,esteban2016predicting}.
For long sequences which require extended knowledge from the past, popular variants of RNN such as Long-Short Term Memory (LSTM) \cite{gers1999learning} and
Gated Recurrent Unit (GRU) \cite{chung2014empirical} have shown remarkable effectiveness in dealing with the vanishing gradients problem
and have been successfully deployed in a number of applications. %Given the close interplay of images and language in vision, it is
%reasonable to expect that exciting new results and developments on analyzing sequential data will continue to emerge in the coming years.
%In fact, even if the data are not purely sequential, from the practical standpoint, one can still %formulate
%and train RNN models which analyze sequentially, leveraging powerful algorithms and libraries that are available today. 

{\em Point estimates, confidence and consequences.} Despite the impressive predictive power of RNN models, the predictions rely on the ``point estimate'' of the parameters.
The confidence score can often be overestimated due to overfitting \cite{fortunato2017bayesian} especially on datasets with insufficient sample sizes.
More importantly, in practice, without acknowledging the level of uncertainty about the prediction, the model
cannot be blindly trusted in mission critical applications. 
%blindly trusting the model output can be
Unexpected performance variations with no sensible way of anticipating this possibility %from a probabilistic perspective
 is also
a limitation in terms of regulatory compliance (e.g., FDA).
When a decision made by a model could result in dangerous outcomes in real-life tasks such as an autonomous vehicle not detecting a pedestrian,
missing a disease prediction due to some artifacts in a medical image, or radiation therapy planning \cite{lambert2011mri}, 
knowing how `certain' the model is about its decision can offer a chance to look for alternative solutions such as alerting the driver to take over
or recommending a different disease test to prevent bad outcomes made by erroneous decisions.

{\em Uncertainty.} When operating with predictions involving data and some model, there are mainly two sources of unpredictability.
First, there may be uncertainty that arises from an inaccurate dataset or observations --- this is called \textit{aleatoric} uncertainty.
On the other hand, the lack of certainty resulting from the model itself (i.e., model parameters) is called \textit{epistemic} uncertainty \cite{der2009aleatory}.
Aleatoric uncertainty comes from the observations {\em externally} such as noise and other factors that cannot typically be inferred systematically,
and so, algorithms instead attempt to calculate {\em the epistemic uncertainty that results from the models themselves}. Hence this is often also referred to as \textit{model uncertainty} \cite{kendall2017uncertainties}.

{\em Related work on uncertainty in Neural networks.} The importance of estimating the uncertainty aspect of neural networks (NN) has been acknowledged in the literature.
%,but there is also consensus that interpreting modern architectures via simple extensions (e.g., sampling output to derive statistical meaning)
%is problematic due to the complex structure of the models we use today. 
%applicable for simpler models immediately becomes inappropriate in terms of accuracy and computation due to the highly complex structure of NNs.
%Instead, old as well as new results in Bayesian neural networks (BNN) attempt to address these issues from a number of different angles.
%In fact, 
Several early ideas investigated a suite of schemes related to Bayesian neural networks (BNN):
Monte Carlo (MC) sampling \cite{mackay1992bayesian}, variational inference \cite{hinton1993keeping} and Laplace approximation \cite{mackay1992practical}.
%but such methods quickly become infeasible as the architecture sizes grow \cite{neal2012bayesian}.
%Consequently, 
More recent works have focused on efficiently approximating posterior distributions to infer predictive uncertainty.
For instance, scalable forms of variational inference approaches \cite{graves2011practical} suggest estimating the evidence lower bound (ELBO)
via Monte Carlo estimation to efficiently approximate the marginal likelihood of the weights.
Similarly, several proposals have extended the variational Bayes approach to perform probabilistic back propagation with assumed density filtering \cite{hernandez2015probabilistic},
 explicitly update the weights of NN in terms of the distribution parameters (i.e., expectation) \cite{blundell2015weight}, or
apply stochastic gradient Langevin dynamics \cite{welling2011bayesian} at large scales.
These methods, however, theoretically rely on the correctness of the prior distribution, which has shown to be crucial for reasonable predictive uncertainties \cite{rasmussen2005healing}
and the strength or validity of the assumption (i.e., mean field independence) for computational benefits.
More recently, an interesting and different perspective on BNN based uncertainty estimated based on Monte Carlo dropout was proposed by Gal et al. \cite{gal2016dropout}
where the authors approximate the predictive uncertainty by using dropout \cite{srivastava2014dropout} at prediction time. 
This approach can be interpreted as an ensemble method where the predictions based on ``multiple networks'' with different dropout
structures \cite{srivastava2014dropout,lakshminarayanan2016simple} yield estimates for uncertainty.
However, while the estimated {\em predictive uncertainty} is less dependent on the data by using a fixed dropout rate independent from the data,
uncertainty estimation on the network parameters (i.e. weights) is naturally compromised since the fixed dropout rates are already imposed on the weights by the algorithm itself.
%{\color{red}the method is fundamentally based on the randomly `sampled' dropout network structures which makes it less appealing for
%  tasks that require fast uncertainty estimations (i.e., detection system of fast moving autonomous vehicles).}
In summary, while the literature is still in a nascent stage, a number of prominent researchers are studying ways in which uncertainty estimates
can be derived for deep neural architectures in a way we have come to expect from traditional statistical analysis. 

{\em Other gaps in our knowledge.} We notice that while the above methods focus on predictive uncertainty, most
strategies do not explicitly attempt to estimate the uncertainty of the entire \textit{intermediate} representations of the network such as neurons, weights, biases and so on.
While such information is understandably less attractive in traditional applications
where our interest mainly lies in the prediction made by the final output layer, the RNN-type sequential NN often utilizes
not only the last layer of neurons but also directly operates on the intermediate neurons for making a sequence of predictions \cite{mikolov2010recurrent}.
Several Bayesian RNNs have been proposed recently \cite{gal2016dropout,lakshminarayanan2016simple,fortunato2017bayesian} but are based on
the BNN models described above. Their deployment is not always feasible under practical time constraints for real-life tasks
especially with high dimensional inputs. Furthermore, empirically more powerful variants of RNNs such as LSTMs or GRUs have not been explicitly studied
in the literature in the context of uncertainty at all.
\begin{figure*}[!t]
	\centering
	$\overbrace{\hspace{0.495\textwidth}}^{\text{Input and Decoder Estimate}}$ $\overbrace{\hspace{0.495\textwidth}}^{\text{Prediction Estimate}}$ \\
	\includegraphics[width=\textwidth]{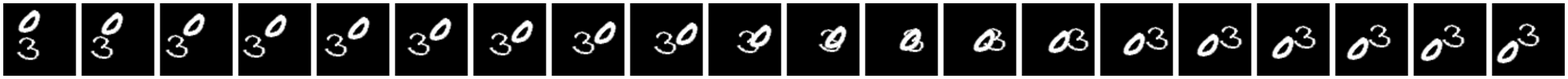} \\ \smallskip
	\includegraphics[width=\textwidth]{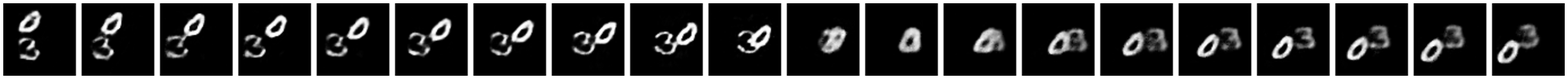} \\ \smallskip
	\includegraphics[width=\textwidth]{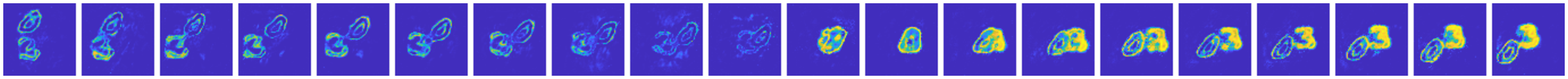} 
	\caption{\label{fig:mnist0} Unsupervised image sequence (moving from left to right) learning with uncertainty using \GRUNAME{}. Top: ground truth input and output. Middle: (left) the reconstruction  of the input  and (right) the prediction  of the output. Bottom: the model's uncertainty map where the bright regions indicate high uncertainty.
%	uncertainty prediction on an image sequence NP-GRU decoder and predictor results. Each row corresponds to the ground truth, mean prediction, and model uncertainty estimate respectively.
}
\end{figure*}

\textbf{Contributions.} Our overarching goal in this paper
is to enable uncertainty estimation on more powerful sequential neural networks, namely gated recurrent units (GRU), while addressing the
issues discussed above in recent developments of BNNs. To our knowledge, few (if any) other works offer this capability. 
We propose a probabilistic GRU, where \textit{all} network parameters follow exponential family distributions.%which is fully parameterized by natural parameters (of certain distributional families). 
We call this framework the \GRUNAME{}, which operates \textit{directly} on these parameters, inspired in part
by an interesting result for non-sequential data \cite{wang2016natural}.
Our \GRUNAME{} directly offers the following properties: 
\begin{inparaenum}[\bfseries (i)]
\item The operations within each cell in the GRU proceed only with respect to the natural parameters \textit{deterministically}. Thus, the overall procedure is completely sampling-free. Such a property is especially appealing for
sequential datasets of small sample size; 
\item Because weights and biases and {\em all intermediate neurons} of \GRUNAME{} can be expressed in terms of a distribution, their uncertainty estimates can be directly inferred from the network itself.
%Uncertainty estimates on the network parameters (i.e., weights and biases)
%and {\em all intermediate neurons} are approximated throughout the propagation process %Since the network is defined entirely in terms of natural parameters, 
%because every variable of \GRUNAME{} can be expressed in
%terms of a distribution.% parameterized by its corresponding natural parameters. 
\item We focus on 
%the natural parameters for some
some well-known exponential family distributions (i.e., Gaussian, Gamma) which have nice characteristics that can be appropriately chosen
with minimal modifications to the operations depending on the application of interest.
\end{inparaenum}

%Until now, we have described methods for uncertainty estimation in traditional networks which have shown to be on varying tasks. Unfortunately, when the focus is on another type of common but highly informative sequential datasets where each sample is a \textit{sequence} of features or inputs, the undesirable properties of the previously mentioned BNNs require even more attention due to the following reasons: (a) MC sampling   data with arbitrarily long sequences of arbitrary

%\textbf{Contributions:}  In this paper, we make the following contributions:
%(1) We describe a probabilistic GRU called natural parameter GRU (NP-GRU) based only on natural parameters of exponential family distributions. (2) probabilistic GRU with expectation propagation-like approach which requires no sampling or variational inference type approximation.

\section{Background and Preliminaries} \label{sec:exp_fam}
We briefly review some concepts that will be useful throughout the paper. 
We denote matrices with uppercase letters and vectors as lowercase.% unless specified otherwise.

The Gated Recurrent Unit (GRU) and the Long-Short Term Memory (LSTM) are popular variants of RNN where the network parameters are shared across layers. While they both deal with exploding/vanishing gradient issues with \textit{cell} structures of similar forms, the GRU specifically does not represent the cell state and hidden state separately. Specifically, its gates and cell/hidden state updates take the following form:
\begin{align}\label{eq:gru}
\textrm{\textbf{Reset Gate:}} \quad r^t &= \sigma(W_r x^t + b_r) \\
\textrm{\textbf{Update Gate:}} \quad z^t &= \sigma(W_z x^t + b_z) \\
\textrm{\textbf{State Candidate:}} \quad \hat{h}^t &= \tanh(U_{\hat{h}} x^t + W_{\hat{h}}(r^t \odot h^{t-1})+ b_{\hat{h}}) \\
\textrm{\textbf{Cell State:}} \quad h^t &= (1-z^t) \odot \hat{h}^t + z^t \odot h^{t-1}
\end{align}
where $W_{\{r,z,\hat{h}\}}$ and $b_{\{r,z,\hat{h}\}}$ are the weights and biases respectively for their corresponding updates.
Typical implementations of both GRUs and LSTMs include an output layer outside of the cell
%that takes as input the current or last hidden state of the GRU and generates the signal desired, whether it be the reproducing the sequential input in an unsupervised auto-encoder \cite{srivastava2015unsupervised} or some target output in the supervised case \cite{chung2014empirical}.
to produce the desired outputs \cite{chung2014empirical,srivastava2015unsupervised,jozefowicz2015empirical}.
%Though LSTMs remain the most common recurrent network architecture used in practice today, the more compact GRU model has been shown to be competitive with LSTMs in many standard machine learning tasks \cite{chung2014empirical,jozefowicz2015empirical}.
However, both models do not naturally admit more than point estimates of hidden states and outputs as with other typical deterministic models.

Moving beyond point estimates requires us to have a solid grasp on the distributions our outputs may take. In classical statistics, the properties of distributions within \textit{exponential families} have been extremely well studied.
\begin{definition}(Exponential Families)\label{def:expfam}
	Let $x \in X$ be a random variable with probability density or mass function (pdf/pmf) $f_X$. Then $f_X$ is an exponential family distribution if 
	\begin{align}\label{eq:exp_fam}
	f_X(x | \eta) = h(x) \exp (\eta^T T(x) - A(\eta))
	\end{align}
	with natural parameters $\eta$, base measure $h(x)$, and sufficient statistics $T(x)$. Constant $A(\eta)$ (log-partition function) ensures the distribution normalizes to 1.
\end{definition}
Common distributions (e.g., Guassian, Bernoulli, Gamma) can be written in this unified `natural form' with specific definitions of $h(x)$, $T(x)$ and $A(\eta)$. For instance, the Gaussian distribution in natural form is given by $\eta = (\alpha,\beta)$, $T(x) = (x,x^2)$ and $h(x) = 1/\sqrt{2\pi}$. The more common parametrization $N(\mu,\sigma^2)$ can be derived by letting $\mu = -\alpha / \beta$ and $\sigma^2= -1/\beta$.

Two key properties of this family of distributions have led to their widespread use: (1) their ability to summarize arbitrary amounts of data $x\sim f_X$ through only their sufficient statistics $T(x)$, and (2) their ability to be efficiently estimated either directly through a closed form maximum likelihood estimator, or through minimizing a convex function with convex constraints.
While Bayesian statistics and classical machine learning have taken complete advantage of these properties,
it is only recently that deep learning has turned its eye toward them.

\subsection{Exponential Families in Networks}
\begin{figure}[!t]
	\centering
	\scalebox{1}{
	\includegraphics[]{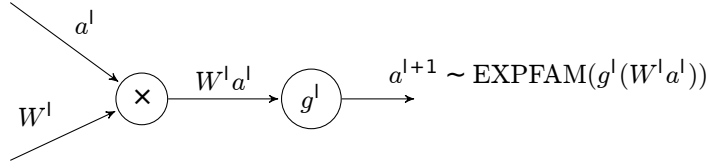}
	}
	\caption{\label{fig:expfam} A single exponential family neuron. Weights $W^l$ are learned, and the output of a neuron is a sample generated from the exponential family defined a priori and by the natural parameters $g^l(W^l a^{l-1})$.}
\end{figure}
Recent work on Deep Exponential Families (DEFs) \cite{ranganath2015deep} explicitly models the output of any given layer as a random variable,
sampled from an exponential family defined by natural parameters given by the linear product of the previous layer's output and a learnable weight matrix (see Fig. \ref{fig:expfam}).
While this formulation leads directly to distributions over hidden states and model outputs, we have not learned distributions over the \textit{model parameters}.
Perhaps even more critical is that we have completely given up computational feasibility: the variational inference procedure used for learning these DEFs requires \textit{Monte Carlo sampling at each hidden state many times for every input sample.}
Indeed, the authors note themselves
%(\href{https://www.youtube.com/watch?v=RQ3dhzGLAvk}{link})
that the cost of running just \textit{text} experiments was \$40K.
In what follows we explore a different line of attack;
one which is \textit{sampling free} and thus allows us to more efficiently and precisely compute the natural parameters of distributions over weights, hidden states, and outputs.

We must first make a key assumption on the distributions of our model parameters. 
In settings where we wish to estimate a random variable weight matrix $W$, we must apply a \textit{mean-field} assumption on its elements $W_{ij}$ such that they are all independent and come from distributions defined by their own individual parameters. For example, let $(\alpha,\beta)$ be the natural parameters of the distribution family chosen. Then the joint distribution over all entries in $W$ is
\begin{equation}\label{eq:mean-field}
p(W|W_\alpha, W_\beta) = \prod_{i,j} p(W(i,j) \ | \ W_\alpha(i,j),W_\beta(i,j))
\end{equation}
where $W_\alpha$ and $W_\beta$ are matrices of $\alpha$ and $\beta$ respectively. This assumption is commonly applied when the dependence among variables of a high-dimensional sample is not expected to hold and/or computationally infeasible to infer.

\section{Sampling-free Probabilistic Networks}\label{sec:npgru}

%\subsection{Natural Parameter Networks}\label{sec:npn}

%We now describe a probabilistic network, the natural parameter network (NPN) \cite{wang2016natural} which fully operates on a set of natural parameters of exponential family distributions. Similar to traditional NNs, the learning process is deterministic yet still captures the probabilistic aspect of the output \textit{and the network itself}, purely as a byproduct of typical NN procedures (i.e., backpropagation).

We now describe a probabilistic network fully operating on a set of natural parameters of exponential family distributions in a \textit{sampling-free} manner.
Inspired by a recent work \cite{wang2016natural}, the learning process, similar to traditional NNs, is deterministic yet still captures the probabilistic aspect of the output \textit{and the network itself}, purely as a byproduct of typical NN procedures (i.e., backpropagation).

%In Sec.~\ref{sec:exp_fam}, we described how we can express a pdf/pmf of an exponential family distribution in terms of a general expression \ref{eq:exp_fam} by appropriately defining its functions (i.e., sufficient statistics $T(x)$).
%\subsection{Exponential Family}
%The exponential family is a well-known set of probability distributions such that a random variable $x \in X$ has a probability density/mass function (pdf/pmf) $f_X$ with a general expression parametrized by natural parameters $\eta$ as
%\begin{equation}\small
%f_X(x | \eta) = h(x) \exp (\eta^T T(x) - A(\eta))
%\end{equation}
%with the following known functions: base measure $h(x)$, sufficient statistics $T(x)$ and log-partition $A(\eta)$.
%By appropriately defining these functions corresponding to a desired exponential family member, many useful distributions (i.e., normal, gamma) can be written exactly in terms of the above `natural form'.
%For instance, given a natural form (which is not unique but often follows a conventional form) of a univariate normal distribution with two natural parameters $\eta=(\alpha,\beta)^T$, we can interpret the distribution with $\eta$ in a more interpretable form  $N(\mu,\sigma^2)$ where $\mu = -\alpha / \beta$ and $\sigma^2= -1/\beta$.
Unlike the probabilistic networks mentioned before, our GRU performs forward propogation in a series of  \textit{deterministic} linear and nonlinear transformations on the distribution of weights and biases. Throughout the entire process, all operations only involve distribution parameters while maintaining their desired distributions after every transformation. For simplicity, we focus on a subset of exponential family distributions with at most two natural parameters $\eta = (\alpha,\beta)^T$ (Gaussian, Gamma and Poisson) which can easily be mapped to familiar forms (i.e., $\mu = -\alpha / \beta$ and $\sigma^2= -1/\beta$ for Gaussian distribution). This can also provide more intuitive interpretation of the network parameters.

%In addition to the versatility of distribution options, characterizing the network by natural parameters provides an intuitive interpretation of the network parameters (weights and biases) and outputs in terms of the distribution chosen to be modeled throughout the network. For instance, if a network is trained assuming Gaussian distribution over weights in terms of $\eta=(\alpha,\beta)^T$, their mean and variance $(\mu,\sigma^2)$ can be directly computed from the natural parameters such that $\mu = -\alpha / \beta$ and $\sigma^2= -1/\beta$ as shown before.

\begin{figure}[!t]
	\centering
	\scalebox{1}{
	\includegraphics[]{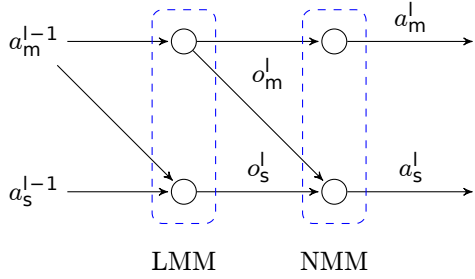}
	}
	\caption{\label{fig:momentmatching} Linear Moment Matching (LMM) is performed at the weights and bias summands, and Nonlinear Moment Matching (NMM) is performed at the sigmoid activation function.}
	
\end{figure}

\subsection{Linear Transformations}\label{sec:linear}
We describe the linear transformation on the input vector $x$ with a matrix $W$ of weights and a vector $b$ of biases in terms of their natural parameters.
We first apply the mean-field assumption on each of the weights and biases based on their individual distribution parameters $\alpha$ and $\beta$ as in Eq. (\ref{eq:mean-field}),
where $\{W_\alpha,W_\beta\}$ and $\{b_\alpha,b_\beta\}$ are the parameters of $W$ and $b$ respectively. Thus, analogous to the linear transformation $o = Wa + b$ in an ordinary neural networks on the previous layer output (or an input) $a$ with $W$ and $b$, the our network operates purely on $(\alpha,\beta)$ to compute $\{o_\alpha,o_\beta\}$.
%: $\{o_\alpha,o_\beta\}$, $\{W_\alpha,W_\beta\}$,  $\{a_\alpha,a_\beta\}$ and $\{b_\alpha,b_\beta\}$.
%Essentially, the NPN network consists of two sets of variables (one for each of the two natural parameters we focus on), and the goal is to perform forward propagation similar to that of a traditional NN.

After each linear transformation, it is necessary to preserve the `distribution property' of the outputs (i.e., $o_\alpha$ and $o_\beta$ still define the same distribution) throughout the forward propagation so that the intermediate nodes and the network itself can be naturally interpreted in terms of their distributions. Thus, we \textit{cannot} simply mimic the typical linear transformation on $a_\beta$ and compute $o_\beta = W_\beta a_\beta + b_\beta$ if we want $o_\beta$ to still be able to preserve the distribution \cite{wang2016natural}.

We perform a second order moment matching on the mean and variance of the distributions. We note that the mean $m$ and variance $s$ can easily be computed with an appropriate function $g(\cdot,\cdot)$ which maps $g:(\alpha,\beta)\rightarrow(\mu,\sigma)$ for each exponential family distribution of our interests (i.e., $g(\alpha,\beta) = (-\frac{\alpha+1}{\beta},\frac{\alpha+1}{\beta^2})$ for a Gamma distribution).
Thus, we compute the $(m,s)$ counterparts of all the $(\alpha,\beta)$-based components (i.e., $(o_m,o_s) = g(o_\alpha, o_\beta)$).
%The mean and variance of the variables we will operate on using the mapping $g(\cdot,\cdot)$ of the desired exponential family distribution are computed first: $(o_m,o_s) = g(o_\alpha, o_\beta)$, $(a_m,a_s) = g(a_\alpha, a_\beta)$, $(W_m,W_s) = g(W_\alpha, W_\beta)$ and $(b_m,b_s) = g(b_\alpha, b_\beta)$.

Now, using the linear output before the activation function, we can now apply Linear Moment Matching (LMM) on (1) the mean $a_m$ following the standard linearity of random variable expectations and (2) the variance $a_s$ as follows:
\begin{equation}\label{eq:o_m}
o_m = W_m a_m + b_m, \quad o_s = W_s a_s + b_s + (W_m \odot W_m)a_s + W_s(a_m \odot a_m)
\end{equation}
%similar to traditional NNs, following the standard linearity of random variable expectations. Next, the variance is matched as follows:
%\begin{equation}\label{eq:o_s}
%o_s = W_s a_s + b_s + (W_m \odot W_m)a_s + W_s(a_m \odot a_m)
%\end{equation}
where $\odot$ is the Hadamard product. Then, we invert back to $(o_\alpha, o_\beta) = g^{-1}(o_m, o_s)$. 
%Though the exact form of the distribution over a specific network parameter or output may not be known,
For the exponential family distributions of our interest involving at most two natural parameters, matching the first two moments (i.e., mean and variance) is a good approximation.
%After matching the mean and variance are inverted back to their corresponding natural parameters using the same mapping function $(o_\alpha, o_\beta) = g^{-1}(o_m, o_s)$.

\subsection{Nonlinear Transformations} \label{sec:nonlinear}

The next key step in NNs is the element-wise nonlinear transformation where we want to apply a nonlinear function $f(\cdot)$ to the linear transformation output output $o$ parametrized by  $\eta=(o_\alpha,o_\beta)$.
This is equivalent to a general random variable transformation given the probability density function (pdf) $p_O$ for $O$ to derive the pdf $p_A$ of $A$ transformed by $a=f(o)$: $p_A(a) = p_O(f(o))|f'(o)|$.
%Since we parametrize $o$ via $\eta=(o_\alpha,o_\beta)$, the transformation operates with respect to the natural parameters in NPN.
\begin{table*}[!t]\label{fig:gru-table}
	
	\caption{\label{table:np-gru} SP-GRU operations in mean and variance. $\odot$ and $[A]^2$ denotes the Hadamard product and $A \odot A$ of a  matrix/vector $A$ respectively. Note the Cell State does not involve nonlinear operations. See Fig.~\ref{fig:np-gru} for the internal cell structure.}
	
  	\centering% \small
  	\setlength{\tabcolsep}{10pt}
  	{\renewcommand{\arraystretch}{1.6}
 		\scalebox{0.95}{
  	\begin{tabular}{lll}
  		\textbf{Operation} & \multicolumn{1}{c}{\begin{tabular}[]{@{}c@{}}\textbf{Linear}\vspace{-5pt}\\ \textbf{Transformation}\end{tabular}}& \begin{tabular}[]{@{}c@{}}\textbf{Nonlinear}\vspace{-5pt}\\ \textbf{Transformation}\end{tabular} \\
%  		 & \textbf{Linear} & \textbf{Nonlinear} \\ 
%		\textbf{Operation} & \textbf{Transformation} & \textbf{Transformation} \\ 		
		\toprule\toprule
  		
  		\textbf{Reset}& 
  		$o_{r,m}^t = U_{r,m} x_m^t + W_{r,m} s_m^{t-1} + b_{r,m}$ 
  		& $r_m^t = \sigma_m(o_{r,m}^t, o_{r,s}^t)$\\ 
%  		$r^t = \sigma(W_r x^t + b_r)$
  		\textbf{Gate}& $o_{r,s}^t = U_{r,s} x_s^t + W_{r,s} h_s^{t-1} + b_{r,s} + [U_{r,m}]^2x_s^t$
  		& $r_s^t = \sigma_s(o_{r,m}^t, o_{r,s}^t)$ \\
  		&\qquad $ + U_{r,s}[x_m^t]^2 + [W_{r,m}]^2h_s^{t-1} + W_{r,s}[h_m^{t-1}]^2$ & \\
  		
  		\hline
  		
  		\textbf{Update} & 
  		$o_{z,m}^t = U_{z,m} x_m^t + W_{z,m} s_m^{t-1} + b_{z,m}$ 
  		& $z_m^t = \sigma_m(o_{z,m}^t, o_{z,s}^t)$\\ 
%  		$z^t = \sigma(W_z x^t + b_z)$
		\textbf{Gate} & $o_{z,s}^t = U_{z,s} x_s^t + W_{z,s} h_s^{t-1} + b_{z,s} + [U_{z,m}]^2x_s^t$
  		& $z_s^t = \sigma_s(o_{z,m}^t, o_{z,s}^t)$ \\
  		&\qquad $ + U_{z,s}[x_m^t]^2 + [W_{z,m}]^2h_s^{t-1} + W_{z,s}[h_m^{t-1}]^2$ & \\
  		\hline
  		
  		\textbf{State} & 
  		$o_{\hat{h},m}^t = U_{\hat{h},m} x_m^t + W_{\hat{h},m} s_m^{t-1} + b_{\hat{h},m} $ 
  		& $\hat{h}_m^t = {\tanh}_m(o_{\hat{h},m}^t, o_{\hat{h},s}^t)$\\ 
%  		$\hat{h}^t \hspace{-1pt}=\hspace{-1pt} \tanh(U_h x^t \hspace{-2pt}+\hspace{-1pt} W_h(r^t \hspace{-1pt}\odot\hspace{-1pt} h^{t-1})\hspace{-1pt}+\hspace{-1pt}b_h)$
  		\textbf{Candidate} & $o_{\hat{h},s}^t = U_{\hat{h},s} x_s^t + W_{\hat{h},s} h_s^{t-1} + b_{\hat{h},s}
  		+ [U_{\hat{h},m}]^2x_s^t $
  		& $\hat{h}_s^t = {\tanh}_s(o_{\hat{h},m}^t, o_{\hat{h},s}^t)$ \\
  		&\qquad $+ U_{\hat{h},s}[x_m^t]^2
  		+ [W_{\hat{h},m}]^2\hat{h}_s^{t-1} + W_{\hat{h},s}[h_m^{t-1}]^2$ & \\
  		\hline
  			
  		\textbf{Cell State} & 
  		$h_m^t =  (1-z_m^t)\odot \hat{h}_m^t + z_m^t \odot h_m^{t-1} $ 
  		& \multirow{2}{*}{Not Needed}\\ 
 % 		$h^t = (1-z^t) \odot \hat{h}^t + z^t \odot h^{t-1}$
  		& $h_s^t = [(1-z_s^t)]^2\odot \hat{h}_m^t + [z_s^t]^2 \odot h_s^{t-1}$ 
  		& \\
  
  		\bottomrule\bottomrule
  	\end{tabular}
  }
  		}
  \end{table*}
  
However, well-known nonlinear functions such as sigmoids and hyperbolic tangents cannot directly be utilized on $(o_\alpha,o_\beta)$ because the resulting $o$ given $\eta$ may not be from the same exponential family distribution. Thus, we perform another second order moment matching in terms of mean $o_m$ and variance $o_s$ via Nonlinear Moment Matching (NMM). Ideally, we need to marginalize over a distribution of $o$ given $(o_\alpha,o_\beta)$ to compute $a_m = \int f(o) p_O(o \ | \ o_\alpha,o_\beta) do$ and the corresponding variance $a_s = \int f(o)^2 p_O(o \ | \ o_\alpha,o_\beta) do - a_m^2$ which we map back to $(a_\alpha,a_\beta)$ with an appropriate bijective mapping function $g(\cdot,\cdot)$.
%Once the mean and variance are obtained, we can simply invert them back to the natural parameters via their corresponding bijective mapping function $g(\cdot,\cdot)$. 
Unfortunately, when the dimension of $o$ grows, which is often the case in NN, the computational burden of integral calculation becomes incredibly more demanding.

The closed form approximations described below can efficiently compute the mean and variance of the activation outputs $a_m$ and $a_s$ \cite{wang2016natural}. We show these approximations for sigmoids $\sigma(x)$ and hyperbolic tangents $\tanh(x)$ for a Gaussian distribution, as these will become the critical components used in our probabilistic GRU. Here, we use the fact that $\sigma(x) \approx \Phi(\zeta x)$ where $\Phi()$ is a probit function and $\zeta = \sqrt{\pi/8}$ is a constant. Then, we can approximate the sigmoid functions for $a_m$ and $a_s$ as
\begin{equation}\small
a_m \approx \sigma_m(o_m, o_s) =  \sigma \left( \frac{o_\alpha}{(1 + \zeta^2 o_\beta)^\frac{1}{2}} \right), \quad a_s \approx \sigma_s(o_m, o_s) = \sigma \left( \frac{\nu(o_m + \omega)}{(1 + \zeta^2 \nu^2o_s)^\frac{1}{2}} \right) - a_m^2
\end{equation}
where $\nu = 4 - 2 \sqrt{2}$ and $\omega = -\log(\sqrt{2}+1)$.
A similar form for the hyperbolic tangent can be derived easily from $\tanh(x) = 2\sigma(2x)-1$. %the hyperbolic tangent function $\tanh_m(\cdot,\cdot)$ for $a_m$ is
%\begin{equation}\small
%a_m \approx \tanh_m(o_m, o_s) = 2\sigma \left( \frac{o_\alpha}{(0.25 + \zeta^2 o_\beta)^\frac{1}{2}} \right) - 1
%\end{equation}
%for $\nu = 8 - 4 \sqrt{2}$ and $\omega = -0.5 \log (\sqrt{2} + 1)$, and $\tanh_s(\cdot,\cdot)$ for $a_s$ is
%\begin{equation}\small
%a_s \approx \tanh_s(o_m, o_s) = 4\sigma \left( \frac{\nu(o_\alpha + \omega)}{(1 + \zeta^2 \nu^2o_\beta)^\frac{1}{2}} \right) - a_m^2 - 2 a_m - 1
%\end{equation}
%where $\nu = 2(4 - 2 \sqrt{2})$ and $\omega = -\log(\sqrt{2}+1) / 2$. 

%While a Gaussian distribution can conveniently construct the closed forms of typical activation functions (including ReLUs), other common exponential family distributions do not have obvious ways to make such straightforward approximations. However, using an `activation-like' mapping $f(x) = a - b \exp (-\gamma d(x))$  where $d(x)$ is an arbitrary activation of choice with appropriate constants $a$, $b$ and $\gamma$, nonlinear transformations of other distributions can be formulated in closed form \cite{wang2016natural}. Derivations for other activation functions and exponential family distributions are included in the supplement.

Note that other common exponential family distributions do not have obvious ways to make such straightforward approximations. Thus, we use an `activation-like' mapping $f(x) = a - b \exp (-\gamma d(x))$  where $d(x)$ is an arbitrary activation of choice with appropriate constants $a$, $b$ and $\gamma$. Nonlinear transformations of other distributions can then be formulated in closed form (see Appendix A for details).%\cite{wang2016natural}. Derivations for other activation functions and exponential family distributions are included in the supplement.

%Note that the approximation errors may be accumulated for deep network structures, such variations can be adjusted with respect to the data during backpropagation \cite{wang2016natural}. 
%\begin{table*}[h]
%	\vspace{-12pt}
%	\centering
%	\setlength{\tabcolsep}{2pt}
%	\scalebox{0.99}{
%		\begin{tabular}{c|l}
%			Gates & Output & \\ \hline \hline
%			Reset & \checkmark & & \\
%			Update & \checkmark & \checkmark & \\
%			Cell State & \checkmark & \checkmark & \checkmark 
%		\end{tabular}
%	}
%	\vspace{-10pt}
%	\caption{\label{table:sg_tasks} \small Scene graph detection tasks. Check marks indicate required prediction components. The tasks become incrementally more demanding from top (PredCls) to bottom (SgGen).}
%	\vspace{-6pt}
%\end{table*}

\subsection{Sampling-free Probabilistic GRU} \label{sec:np-gru}
\begin{figure}[t]
	\centering
\includegraphics[width=0.8\columnwidth]{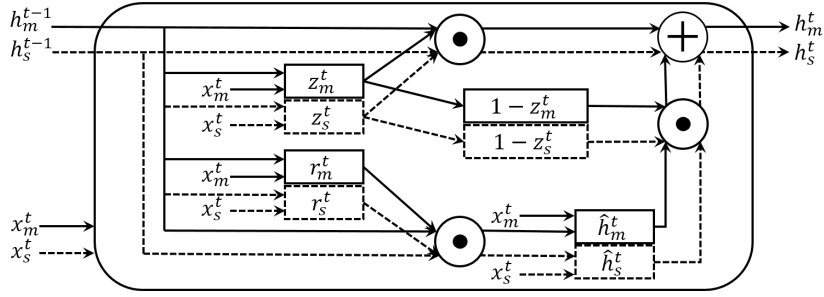}
\caption{\footnotesize\label{fig:np-gru} \GRUNAME{} cell structure. Black lines/boxes and red dotted lines/boxes correspond to operations with respect to variables in mean $m$ and variance $s$ respectively. Circles denote element-wise operators. Note that even though it is illustrated in $m$ and $s$ to relate this to the list of \GRUNAME{} operations in Table~\ref{table:np-gru}, the structure can be described in terms of their correspondingly mapped natural parameters $\alpha$ and $\beta$.}
\end{figure}

Based on the probabilistic formulations described above, we present Sampling-free Probabilistic GRU, the (\GRUNAME{}). The internal architecture is shown in Fig.~\ref{fig:np-gru}. Here, we focus on adapting GRU with the Sampling-free Probabilistic (SP) formulation.
% Let us first briefly define the GRU architecture we focus on which consists of the following: (1) Reset gate $r^t = \sigma(W_r x^t + b_r)$, (2) Update gate $z^t = \sigma(W_z x^t + b_z)$, (3) State candidate $\hat{h}^t = \tanh(U_h x^t + W_h(r^t \odot h^{t-1}) + b_h)$ and (4) Hidden state $h^t = (1-z^t) \odot \hat{h}^t + z^t \cdot h^{t-1}$ where $W$ and $b$ are corresponding weights and biases and $x^t$ is an input vector at step $t$.

We express all the variables related to the GRU (left column of Table~\ref{table:np-gru}) in terms of their parameters $\eta = (\alpha,\beta)$. Specifically, each of the 
variables that is related to a certain operation ($r$, $z$, $\hat{h}$ or $h$) has an additional subscript indicating its associated gate. For instance, $W_r$ is now expressed \textit{only} in terms of its parameters $W_{r,\alpha}$ and $W_{r,\beta}$ (i.e., two weight matrices). Again, we assume that all of the variables are factorized. Notice that since the GRU consists of a series of operations (i.e., gates) where each operation is a NN with linear and nonlinear transformations, we can update each gate by the transformations defined above.%treat each gate as its own natural parameter neuron and define the transformations as above.

Assuming that the desired exponential family distribution provides an invertible parameter mapping function $g(\cdot, \cdot)$, we first transform all of the natural parameter variables to means and variances.
Then, given an input sequence $x = \{x_m^1,x_s^1\},\dots, \{x_m^T,x_s^T\}$, we perform linear/nonlinear transformations with respect to means and variances for each GRU operation (Fig.~\ref{fig:np-gru}, Table \ref{table:np-gru}).

The cell state computation does not involve a nonlinear transformation. For an output layer on the hidden states to compute the desired estimate $\hat{y}$, a typical  layer can be defined in a similar manner to obtain both $\hat{y}_m$ and $\hat{y}_s$. In the experiments that we describe next, we add another such layer to compute the mean and variance of the sequence of predictions $\hat{y} = \{y_m^1,y_s^1\},\{y_m^2,y_s^2\}, \dots, \{y_m^T,y_s^T\}$.

%{\color{red} motivation/justification of NP-GRU} 
{\em Extensibility remarks.} Here, we note that despite the simplicity of the cell structure of the \GRUNAME{} as shown in Fig.~\ref{fig:np-gru}, where the flow of the internal pipeline roughly resembles a standard GRU setup \cite{chung2014empirical}, our exponential family adaptation is \textit{not} limited to GRU. For instance, the above formulation can be extended to other variants of RNNs such as LSTMs. Its additional gates and cells are still sigmoid or hyperbolic tangent functions. With the rapidly growing appearance of many RNN variants \cite{greff2017lstm,athiwaratkun2017malware}, 
a versatile probabilistic RNN-type model which is not only empirically competitive but also of sound interpretability is of high value in the immediate future. 
%Second, 

%Naturally, in the GRU construction which receives a series of inputs at each step, the approximation error is conveniently regulated throughout the forward propagation as well.

\section{Experiments}
%We evaluate our model on two different applications involving imaging data. 
%First, 
We perform unsupervised learning of decoding and predicting image sequences from the moving 
MNIST dataset \cite{srivastava2015unsupervised}.
%Second, we apply our model to a unique neuroscience dataset, 
%consisting of brain imaging acquisitions from individuals at risk for developing Alzheimer's disease.
%\input{fig_angle_var.tex}
All experiments were run on an NVIDIA GeForce GTX 1080 TI graphics card using TensorFlow, with learning rate $\alpha = 0.05$. ADAM optimization \cite{kingma2014adam} was used with an exponential decay rate of $\beta_1 = 0.9$, $\beta_2 = 0.999$ for the first and second moments, respectively. We use the Gaussian exponential family for all setups.
%Extensive technical and experimental details in both settings can be found in the supplement.
\begin{figure}[!h]
	\centering
	%	\vspace{-10pt}
	\includegraphics[width=0.335\columnwidth]{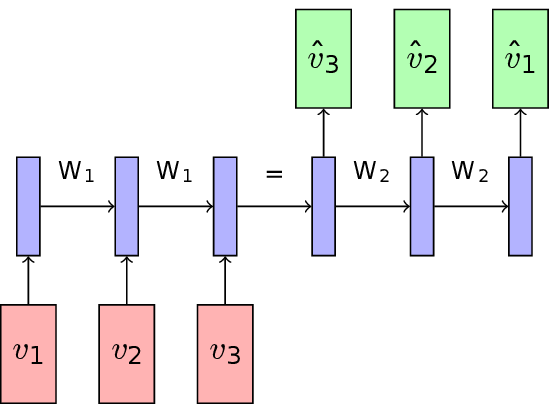}
	\hspace{20pt}
	\includegraphics[width=0.335\columnwidth]{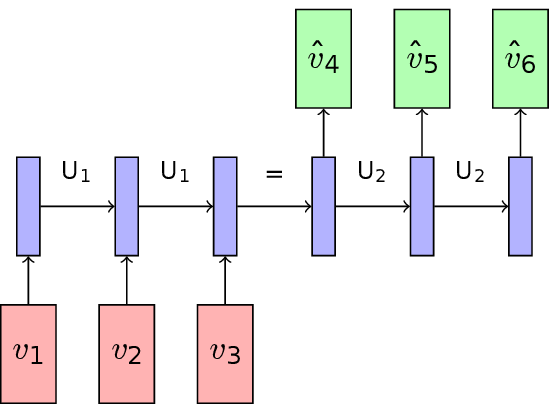}
	\caption{\label{fig:model} Given an input sequence $v_1$,$v_2$,$v_3$, (Left) Autoencoder outputs $\hat{v_3}$,$\hat{v_2}$,$\hat{v_1}$, and (Right) Predictor Network outputs $\hat{v_4}$,$\hat{v_5}$,$\hat{v_6}$.}
\end{figure}

\subsection{Moving MNIST for Unsupervised Sequence Learning}
%\noindent\textbf{Dataset.} 
%Training: two randomly generated MNIST digits bouncing around for $20$ frames. There are two test sets: (1) Same domain: fixed $N=10000$ test sequences with two digits bouncing around for $20$ frames and (2) Unfamiliar domain: randomly generated sequences of only one digit or three digits floating around for $20$ frames.
\noindent\textbf{Goal.} For pixel-level tasks, prediction quality can be understood by the uncertainty estimate, i.e., estimated model variance of that pixel.
In these experiments, we ask the following questions qualitatively and quantitatively:
(1) Given a visually `good looking' sequence prediction, how can we tell that its trajectory is correct?
(2) If it is, can we derive the degree of uncertainty on its prediction?

%Specifically, we can ask the following questions: (1) Given a sensible prediction, how can we tell `correctness', or certainty (i.e., visually sensible but incorrect output)?
%(2) Also, can we derive the degree of uncertainty, i.e., more uncertain about unfamiliar observations?
%can we infer by how much the model is uncertain about its prediction, i.e., more uncertain about unfamiliar observations?
%However, assessing pixel-level uncertainty of image sequences is not a trivial task since the `ground truth uncertainty' references are not readily available.
%From the following experiments, we make both qualitative and quantitative evaluations of uncertainty in various settings to demonstrate its use and capacity.

\noindent\textbf{Experimental Setup.} The moving MNIST dataset consists of MNIST digits moving within a $64 \times 64$ image. A sequence of 20 frames is constructed for each training sample, where digits move randomly, bouncing when a frame edge is hit. Here, the training set consists of an infinite number of potential sequences, and at each iteration a batch of these samples is used.
%\begin{wrapfigure}[15]{r}{0.35\textwidth}
%	%	\vspace{-20pt}
%	\centering
%	\includegraphics[width=0.33\textwidth]{trajs/traj_angle.png}
%	%	\vspace{-5pt}
%	\caption{\label{fig:traj} Controlled moving MNIST trajectories within a frame.}
%	%	\vspace{-20pt}
%\end{wrapfigure}
Following previous work on image sequences \cite{srivastava2015unsupervised}, we split sequences into two parts and evaluate \GRUNAME{} in two similar but distinct network setups. First, we train an \GRUNAME{} auto-encoder, which given the first half of a sequence, learns a hidden representation to be used in the reconstruction of that same sequence (see Fig.~\ref{fig:model}). Second, we set up a predictor \GRUNAME{}, which learns to extrapolate the latter half of the sequence from that hidden representation. All hidden representations are of size $2048$, with $1024$ parameters for each of the two natural parameters.

\begin{figure*}[!h]
	\centering
		\begin{minipage}{110pt} \centering
		\includegraphics[width=0.97\textwidth]{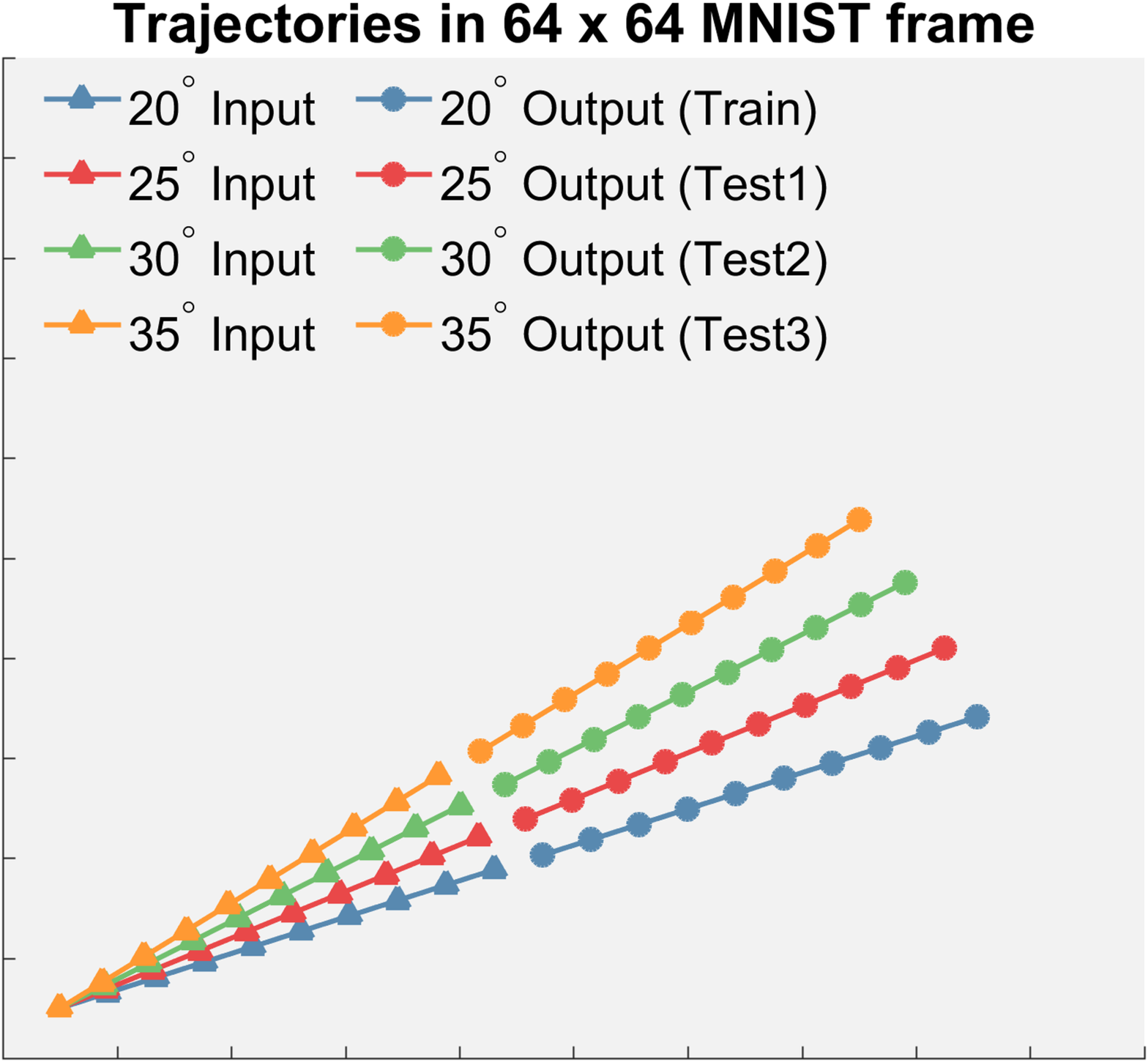}
		(a) Angles
		\end{minipage}
		\begin{minipage}{280pt} \centering
		\includegraphics[width=0.47\textwidth]{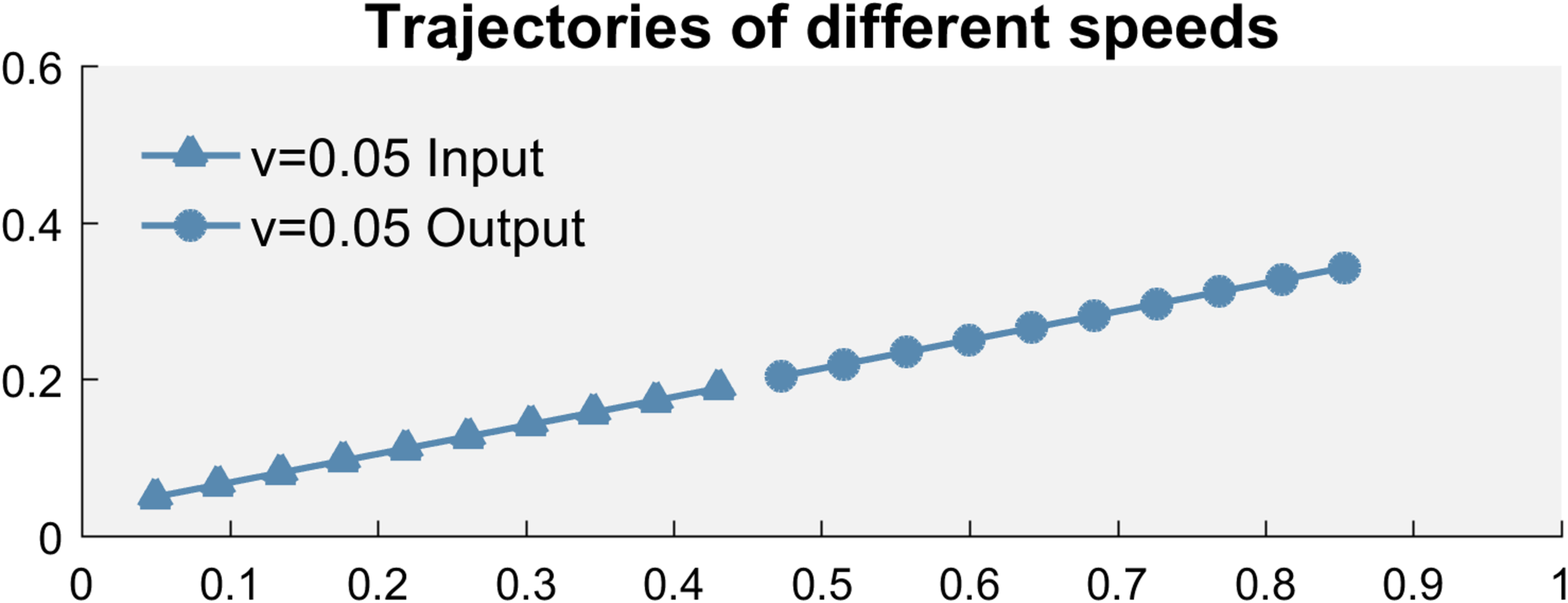}
		\includegraphics[width=0.47\textwidth]{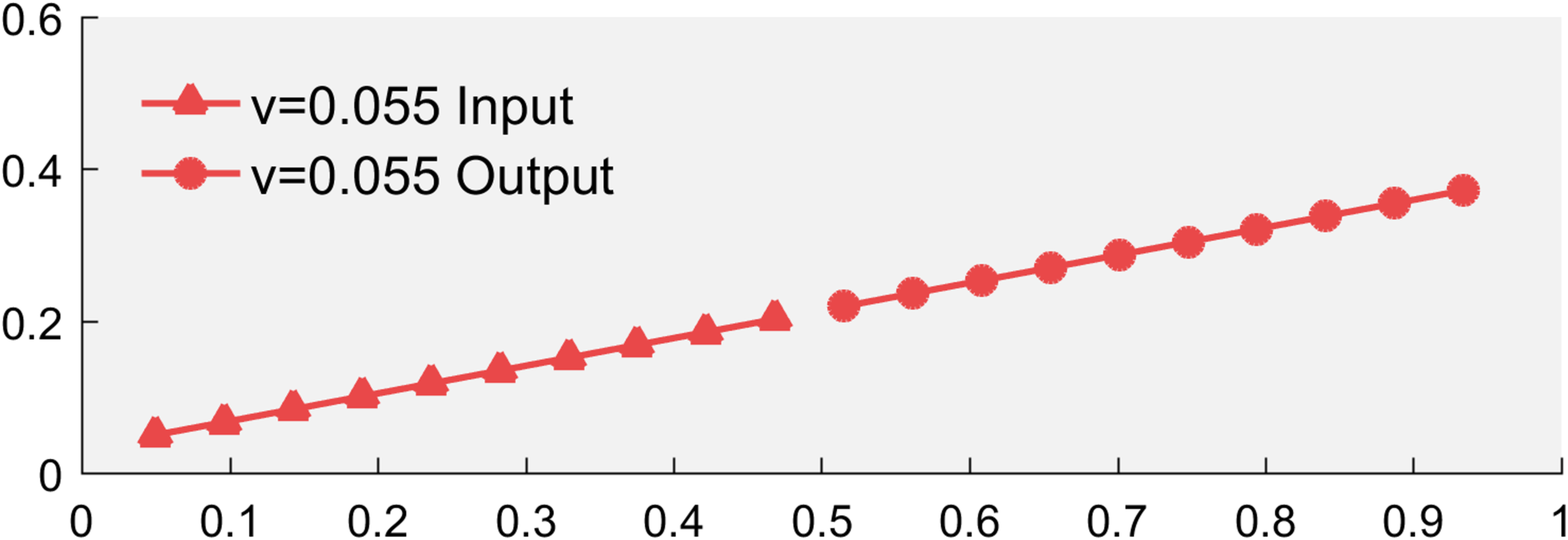}\\
		\includegraphics[width=0.47\textwidth]{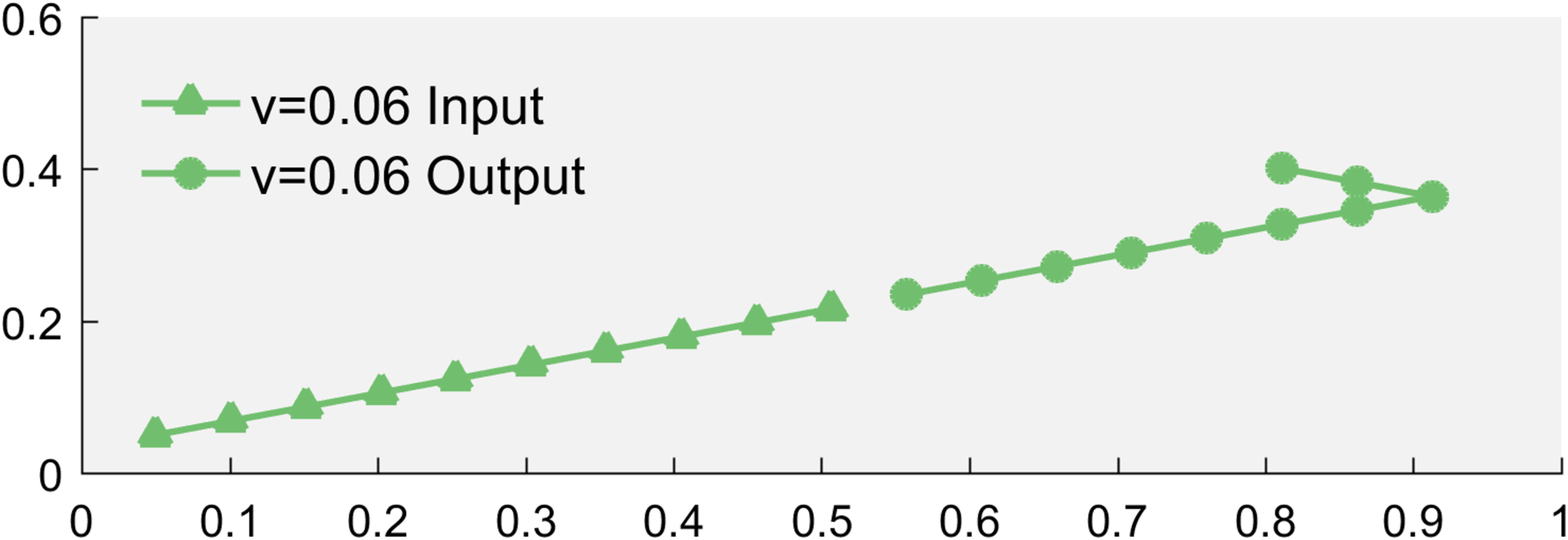}
		\includegraphics[width=0.47\textwidth]{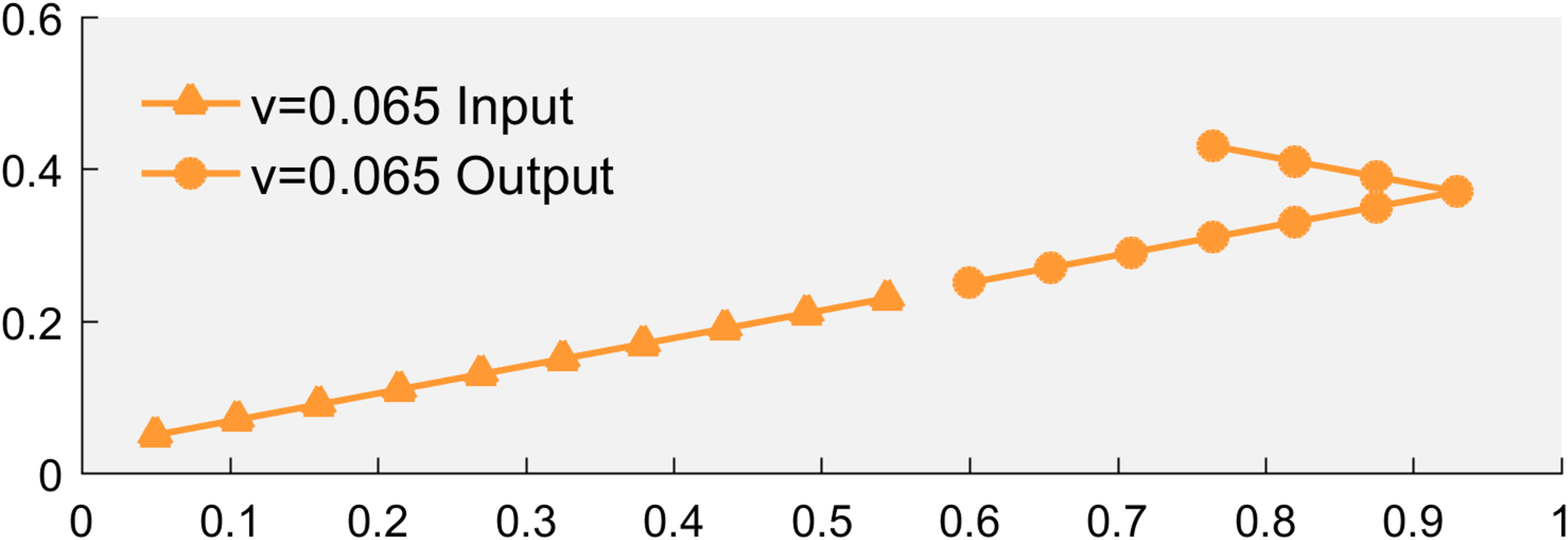}
		\\
		(b) Speeds
		\end{minipage}
		\caption{\label{fig:traj} (a) Angle deviation trajectories. (b) Speed deviation trajectories.}
\end{figure*}

\noindent\textbf{Controlled Moving MNIST.}
%In this controlled experiment, we (1) first train a model for predicting the latter half of moving MNISTs (one digit) following a \textit{specific training path} and (2) test on moving MNISTs with \textit{unfamiliar paths} to intentionally induce uncertainty.
We first train our \GRUNAME{} and the MC-LSTM \cite{gal2016dropout} with the same number of parameters until they have similar test errors (independent of uncertainty) on simple one-digit MNIST sequences moving in a straight line (blue line in Fig.~\ref{fig:traj}).
We then construct three sets of $100$ `unfamiliar' samples where each set consists of sequences deviating from the training sequence path with varying angle, speed and pixel-level noise.

\begin{enumerate}
\item{Angle (Fig.~\ref{fig:traj}a):} We show the input (first 10 triangles) and output (last 10 circles) trajectories of training angle $20^\circ$ which we train the model with. Note that this is the only path that the single digit MNIST sequences traveled. The test trajectories with the angles $25^\circ$, $30^\circ$ and $35^\circ$ which become more `unfamiliar' as it increases are tested to predict the last 10 frames (circles).
\item{Speed (Fig.~\ref{fig:traj}b):} We keep the angle (same training angle as the angle deviation setup show in Fig.~\ref{fig:traj}a) and the speed ($5\%$ of image size per frame) as in top left of Fig.~\ref{fig:traj}b in training. Then, we increase the speed (Fig.~\ref{fig:traj}b: top right, bottom left, bottom right) to $5.5\%$, $6\%$ and $6.5\%$. Thus, the trajectories all travel in the same direction but with different speeds to construct paths with unfamiliar speeds.
\item{Pixel-level noise:} This is a simple setup where the training and testing paths all have the same angles and speeds, but pixel-level noise is added with higher intensities for testing sequences. Specifically, for each pixel a noise from Unif$(0,b)$ is added so the training has zero noise with Unif$(0,b)$ for $b=0$ and the testing sequences have noise with Unif$(0,b)$ for increasing $b=0.2, 0.4, 0.6$.
\end{enumerate}
%We used sampling multiple predictions with different drop-out configurations during the \textit{prediction time}.
%Then, we test on a trajectories deviating from the train samples in the : angle, speed and pixel-level noise.
%Then, we setup three separate experiments to impose various types of deviations to the train trajectory: angle, speed and pixel-level noise.
%In all cases, single SP-GRU and MC-LSTM models were trained based on a set of randomly generated simple one-digit MNIST moving in a straight line (blue line in Fig.~\ref{fig:traj}) and tested in 100 test samples.

\begin{figure*}[!t]
	\centering
	\captionsetup[subfigure]{justification=centering,farskip=5pt,captionskip=1pt}
	\setlength{\fboxsep}{0pt}
	\begin{minipage}{340pt} \centering
	%------------------------------------------------------------------------------- ANGLE
		\begin{minipage}{200pt}
			\begin{minipage}{195pt} \centering
				\hspace{-2pt}
				\begin{minipage}{6pt} \centering
					\vspace{2pt}\textsf{\footnotesize $\theta$} 
				\end{minipage}
				\hspace{2pt}
				\begin{minipage}{53pt} \centering
					\vspace{2pt}\textsf{\footnotesize Ground Truth} 
				\end{minipage}
				\hspace{-2pt}
				\begin{minipage}{53pt} \centering
					\vspace{2pt}\textsf{\footnotesize Prediction} 
				\end{minipage}
				\hspace{-2pt}
				\begin{minipage}{53pt} \centering
					\vspace{2pt}\textsf{\footnotesize Uncertainty} 
					\vspace{-2pt}
				\end{minipage}
				\hspace{-15pt}
			\end{minipage}\\ 
			\fcolorbox{c1}{white}{\setlength{\fboxrule}{1pt}
				\begin{minipage}{190pt} \centering
					\hspace{-10pt}
					$20^{\circ}$
					\hspace{-8pt}
					\begin{minipage}{60pt} \centering
						\includegraphics[width=0.9\textwidth]{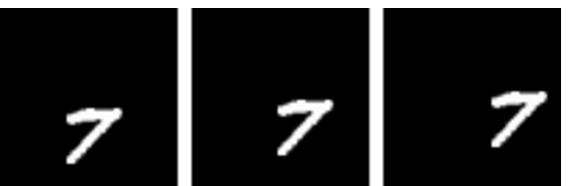}
					\end{minipage}
					\hspace{-9pt}
					\begin{minipage}{60pt} \centering 
						\includegraphics[width=0.9\textwidth]{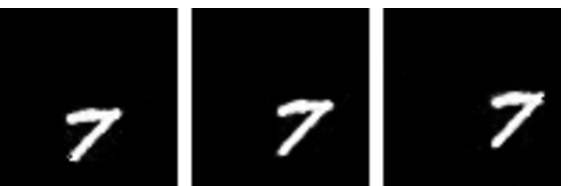}
					\end{minipage}
					\hspace{-9pt}
					\begin{minipage}{60pt} \centering
						\includegraphics[width=0.9\textwidth]{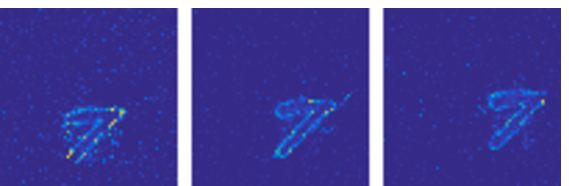}
					\end{minipage}
					\hspace{-15pt}
				\end{minipage}
			}\\
			\fcolorbox{c2}{white}{\setlength{\fboxrule}{1pt}
				\begin{minipage}{190pt} \centering
					\hspace{-10pt}
					$25^{\circ}$
					\hspace{-8pt}
					\begin{minipage}{60pt} \centering
						\includegraphics[width=0.9\textwidth]{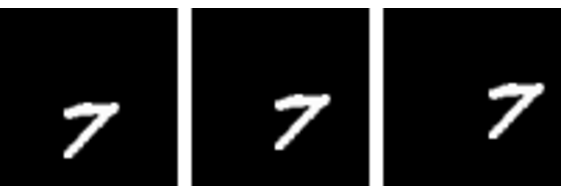}
					\end{minipage}
					\hspace{-9pt}
					\begin{minipage}{60pt} \centering 
						\includegraphics[width=0.9\textwidth]{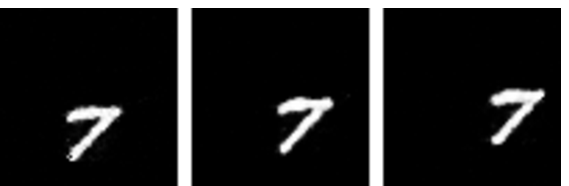}
					\end{minipage}
					\hspace{-9pt}
					\begin{minipage}{60pt} \centering
						\includegraphics[width=0.9\textwidth]{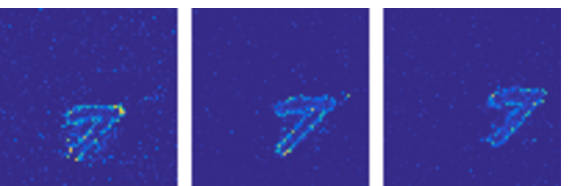}
					\end{minipage}
					\hspace{-15pt}
				\end{minipage}
			}\\
			\fcolorbox{c3}{white}{\setlength{\fboxrule}{1pt}
				\begin{minipage}{190pt} \centering
					\hspace{-10pt}
					$30^{\circ}$
					\hspace{-8pt}
					\begin{minipage}{60pt} \centering
						\includegraphics[width=0.9\textwidth]{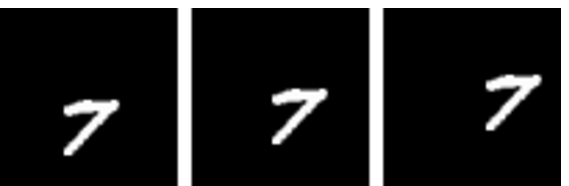}
					\end{minipage}
					\hspace{-9pt}
					\begin{minipage}{60pt} \centering 
						\includegraphics[width=0.9\textwidth]{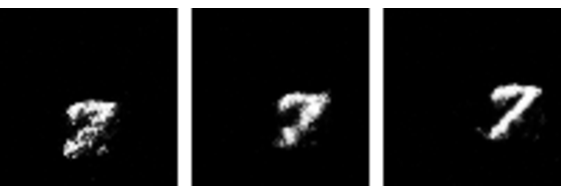}
					\end{minipage}
					\hspace{-9pt}
					\begin{minipage}{60pt} \centering
						\includegraphics[width=0.9\textwidth]{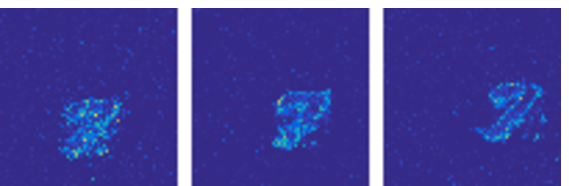}
					\end{minipage}
					\hspace{-15pt}
				\end{minipage}
			}\\
			\fcolorbox{c4}{white}{\setlength{\fboxrule}{1pt}
				\begin{minipage}{190pt} \centering
					\hspace{-10pt}
					$35^{\circ}$
					\hspace{-8pt}
					\begin{minipage}{60pt} \centering
						\includegraphics[width=0.9\textwidth]{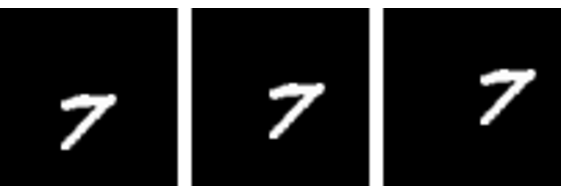}
					\end{minipage}
					\hspace{-9pt}
					\begin{minipage}{60pt} \centering 
						\includegraphics[width=0.9\textwidth]{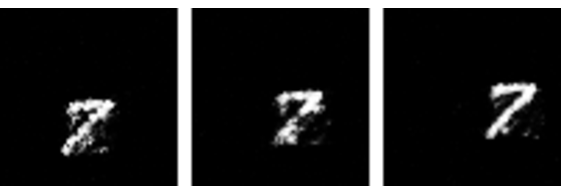}
					\end{minipage}
					\hspace{-9pt}
					\begin{minipage}{60pt} \centering
						\includegraphics[width=0.9\textwidth]{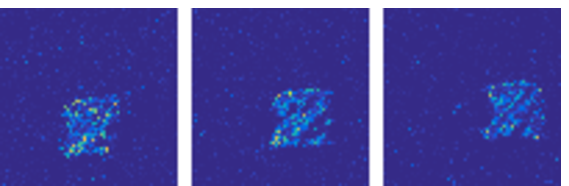}
					\end{minipage}
					\hspace{-15pt}
				\end{minipage}
			}\\
		\end{minipage}
		\hspace{10pt}
		\begin{minipage}{120pt} \centering
		\vspace{5pt}
			\includegraphics[width=\textwidth]{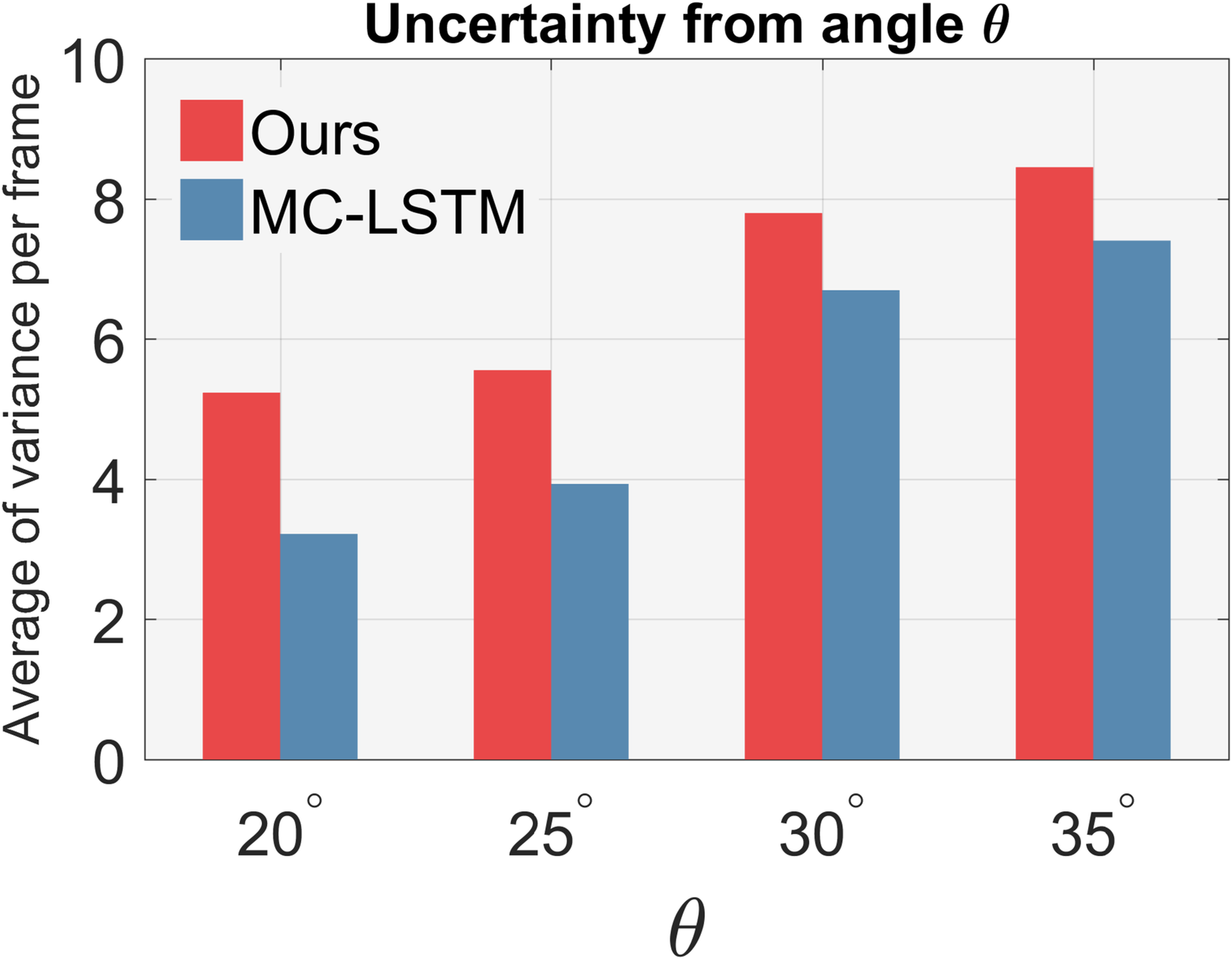}
		\end{minipage}
		
			%------------------------------------------------------------------------------- SPEED
		\begin{minipage}{200pt}
				\begin{minipage}{185pt} \centering
					\hspace{-2pt}
					\begin{minipage}{6pt} \centering
						\vspace{2pt}\textsf{\footnotesize $v$} 
					\end{minipage}
					\hspace{2pt}
					\begin{minipage}{53pt} \centering
						\vspace{2pt}\textsf{\footnotesize Ground Truth} 
					\end{minipage}
					\hspace{-2pt}
					\begin{minipage}{53pt} \centering
						\vspace{2pt}\textsf{\footnotesize Prediction} 
					\end{minipage}
					\hspace{-2pt}
					\begin{minipage}{53pt} \centering
						\vspace{2pt}\textsf{\footnotesize Uncertainty} 
						\vspace{-2pt}
					\end{minipage}
					\hspace{-20pt}
				\end{minipage}\\
				\fcolorbox{c1}{white}{\setlength{\fboxrule}{1pt}
					\begin{minipage}{190pt} \centering
						\hspace{-10pt}
						{\scriptsize $5.0$\%}
						\hspace{-8pt}
						\begin{minipage}{60pt} \centering
							\includegraphics[width=0.9\textwidth]{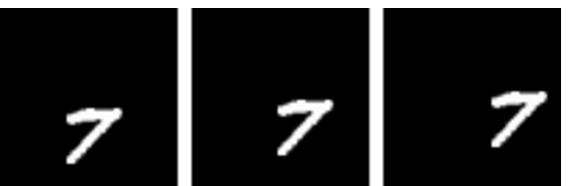}
						\end{minipage}
						\hspace{-9pt}
						\begin{minipage}{60pt} \centering 
							\includegraphics[width=0.9\textwidth]{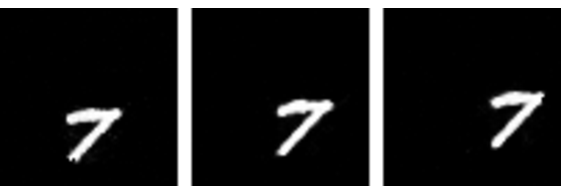}
						\end{minipage}
						\hspace{-9pt}
						\begin{minipage}{60pt} \centering
							\includegraphics[width=0.9\textwidth]{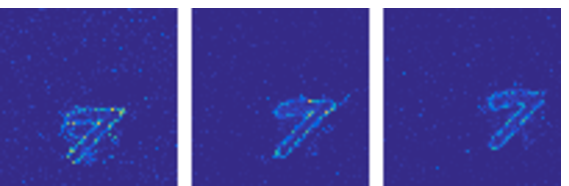}
						\end{minipage}
						\hspace{-10pt}
					\end{minipage}
				}\\
				\fcolorbox{c2}{white}{\setlength{\fboxrule}{1pt}
					\begin{minipage}{190pt} \centering
						\hspace{-10pt}
						{\scriptsize  $5.5$\%}
						\hspace{-8pt}
						\begin{minipage}{60pt} \centering
							\includegraphics[width=0.9\textwidth]{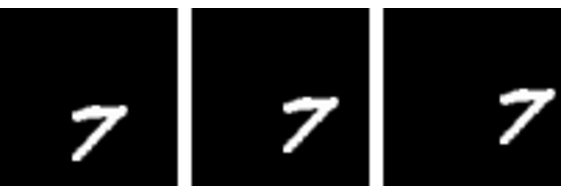}
						\end{minipage}
						\hspace{-9pt}
						\begin{minipage}{60pt} \centering 
							\includegraphics[width=0.9\textwidth]{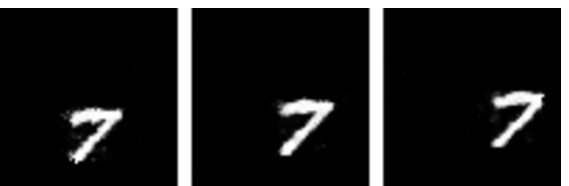}
						\end{minipage}
						\hspace{-9pt}
						\begin{minipage}{60pt} \centering
							\includegraphics[width=0.9\textwidth]{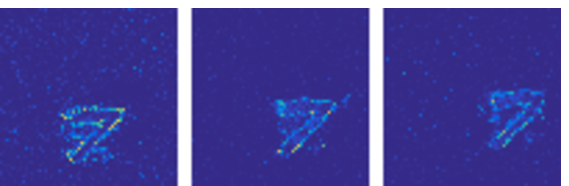}
						\end{minipage}
						\hspace{-10pt}
					\end{minipage}
				}\\
				\fcolorbox{c3}{white}{\setlength{\fboxrule}{1pt}
					\begin{minipage}{190pt} \centering
						\hspace{-10pt}
						{\scriptsize  $6.0$\%}
						\hspace{-8pt}
						\begin{minipage}{60pt} \centering
							\includegraphics[width=0.9\textwidth]{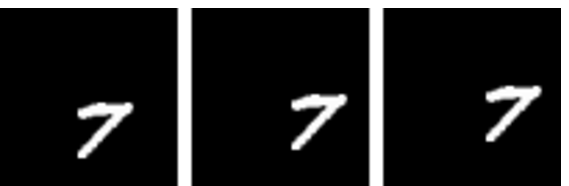}
						\end{minipage}
						\hspace{-9pt}
						\begin{minipage}{60pt} \centering 
							\includegraphics[width=0.9\textwidth]{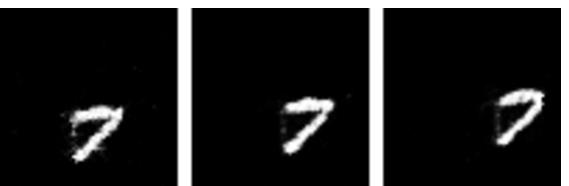}
						\end{minipage}
						\hspace{-9pt}
						\begin{minipage}{60pt} \centering
							\includegraphics[width=0.9\textwidth]{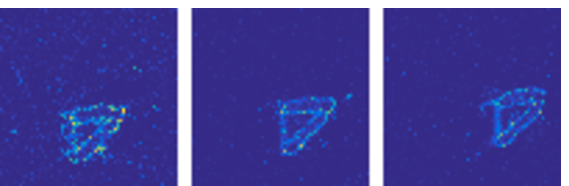}
						\end{minipage}
						\hspace{-10pt}
					\end{minipage}
				}\\
				\fcolorbox{c4}{white}{\setlength{\fboxrule}{1pt}
					\begin{minipage}{190pt} \centering
						\hspace{-10pt}
						{\scriptsize  $6.5$\%}
						\hspace{-8pt}
						\begin{minipage}{60pt} \centering
							\includegraphics[width=0.9\textwidth]{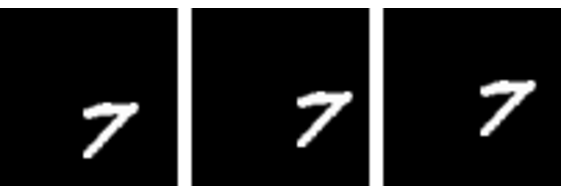}
						\end{minipage}
						\hspace{-9pt}
						\begin{minipage}{60pt} \centering 
							\includegraphics[width=0.9\textwidth]{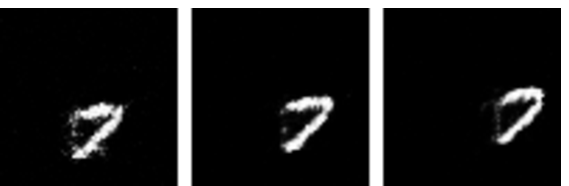}
						\end{minipage}
						\hspace{-9pt}
						\begin{minipage}{60pt} \centering
							\includegraphics[width=0.9\textwidth]{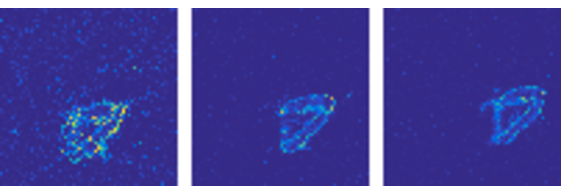}
						\end{minipage}
						\hspace{-10pt}
					\end{minipage}
				}\\
			\end{minipage}
			\hspace{10pt}
			\begin{minipage}{120pt} \centering
					\vspace{5pt}
				\includegraphics[width=\textwidth]{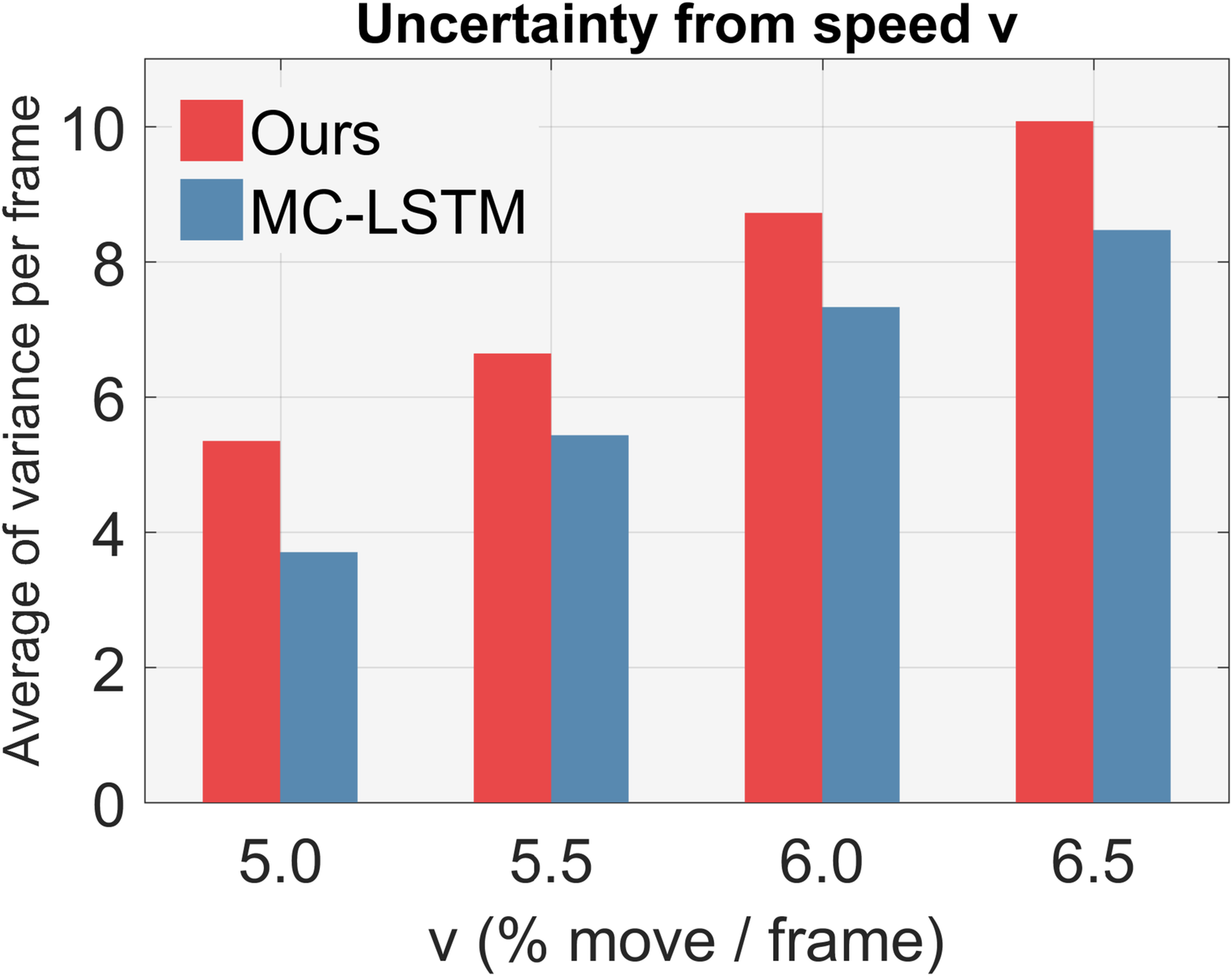}
			\end{minipage}
			
				%------------------------------------------------------------------------------- NOISE
			\begin{minipage}{200pt}
					\begin{minipage}{185pt} \centering
						\hspace{-2pt}
						\begin{minipage}{6pt} \centering
							\vspace{2pt}\textsf{\footnotesize $b$} 
						\end{minipage}
						\hspace{2pt}
						\begin{minipage}{53pt} \centering
							\vspace{2pt}\textsf{\footnotesize Input} 
						\end{minipage}
						\hspace{-2pt}
						\begin{minipage}{53pt} \centering
							\vspace{2pt}\textsf{\footnotesize Prediction} 
						\end{minipage}
						\hspace{-2pt}
						\begin{minipage}{53pt} \centering
							\vspace{2pt}\textsf{\footnotesize Uncertainty} 
							\vspace{-2pt}
						\end{minipage}
						\hspace{-15pt}
					\end{minipage}\\
					\fcolorbox{c1}{white}{\setlength{\fboxrule}{1pt}
						\begin{minipage}{190pt} \centering
							\hspace{-15pt}
							{\footnotesize $0.0$}
							\hspace{-3pt}
							\begin{minipage}{60pt} \centering
								\includegraphics[width=0.9\textwidth]{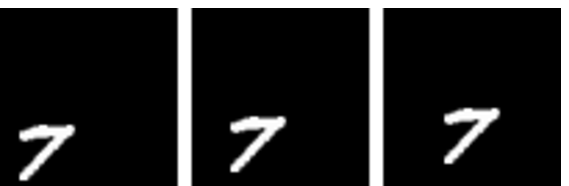}
							\end{minipage}
							\hspace{-9pt}
							\begin{minipage}{60pt} \centering 
								\includegraphics[width=0.9\textwidth]{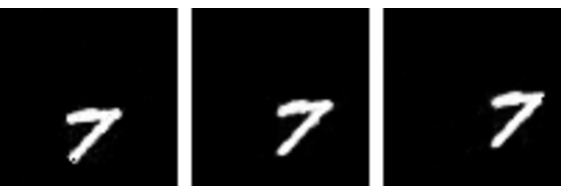}
							\end{minipage}
							\hspace{-9pt}
							\begin{minipage}{60pt} \centering
								\includegraphics[width=0.9\textwidth]{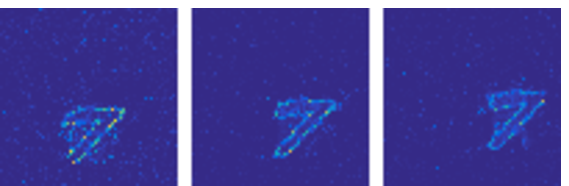}
							\end{minipage}
							\hspace{-15pt}
						\end{minipage}
					}\\
					\fcolorbox{c2}{white}{\setlength{\fboxrule}{1pt}
						\begin{minipage}{190pt} \centering
							\hspace{-15pt}
							{\footnotesize $0.2$}
							\hspace{-3pt}
							\begin{minipage}{60pt} \centering
								\includegraphics[width=0.9\textwidth]{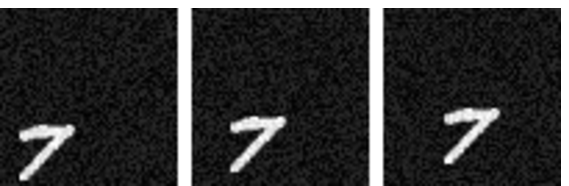}
							\end{minipage}
							\hspace{-9pt}
							\begin{minipage}{60pt} \centering 
								\includegraphics[width=0.9\textwidth]{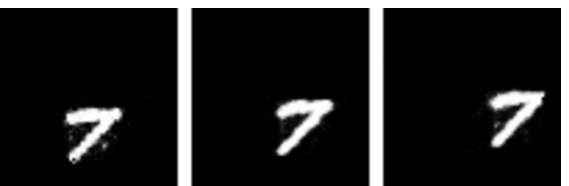}
							\end{minipage}
							\hspace{-9pt}
							\begin{minipage}{60pt} \centering
								\includegraphics[width=0.9\textwidth]{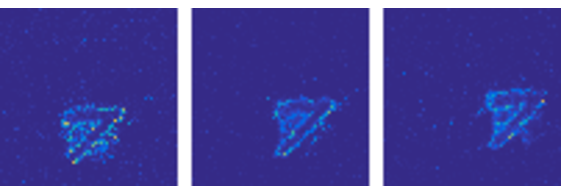}
							\end{minipage}
							\hspace{-15pt}
						\end{minipage}
					}\\
					\fcolorbox{c3}{white}{\setlength{\fboxrule}{1pt}
						\begin{minipage}{190pt} \centering
							\hspace{-15pt}
							{\footnotesize $0.4$}
							\hspace{-3pt}
							\begin{minipage}{60pt} \centering
								\includegraphics[width=0.9\textwidth]{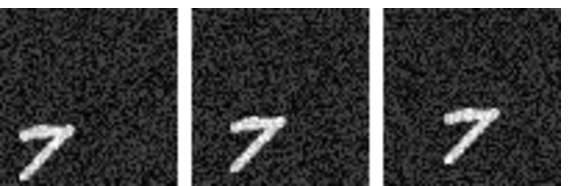}
							\end{minipage}
							\hspace{-9pt}
							\begin{minipage}{60pt} \centering 
								\includegraphics[width=0.9\textwidth]{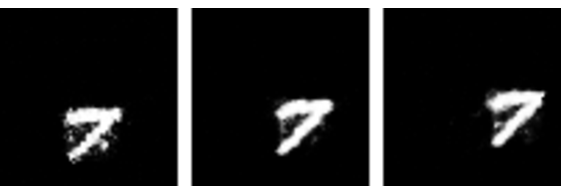}
							\end{minipage}
							\hspace{-9pt}
							\begin{minipage}{60pt} \centering
								\includegraphics[width=0.9\textwidth]{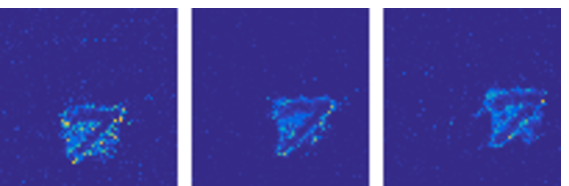}
							\end{minipage}
							\hspace{-15pt}
						\end{minipage}
					}\\
					\fcolorbox{c4}{white}{\setlength{\fboxrule}{1pt}
						\begin{minipage}{190pt} \centering
							\hspace{-15pt}
							{\footnotesize $0.6$}
							\hspace{-3pt}
							\begin{minipage}{60pt} \centering
								\includegraphics[width=0.9\textwidth]{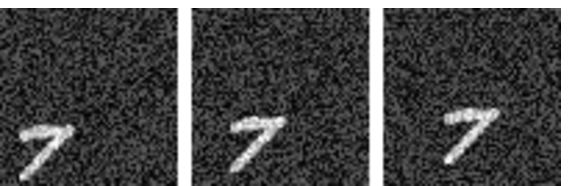}
							\end{minipage}
							\hspace{-9pt}
							\begin{minipage}{60pt} \centering 
								\includegraphics[width=0.9\textwidth]{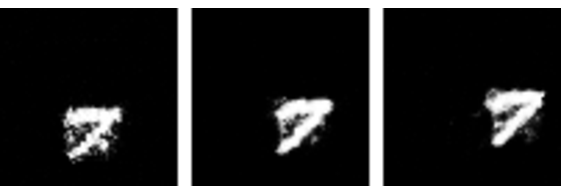}
							\end{minipage}
							\hspace{-9pt}
							\begin{minipage}{60pt} \centering
								\includegraphics[width=0.9\textwidth]{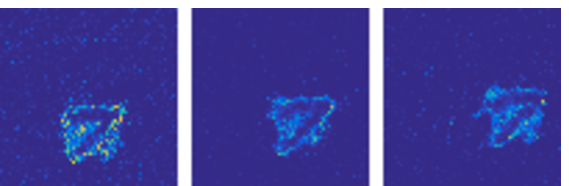}
							\end{minipage}
							\hspace{-15pt}
						\end{minipage}
					}\\
				\end{minipage}
				\hspace{10pt}
				\begin{minipage}{120pt} \centering
						\vspace{5pt}
					\includegraphics[width=\textwidth]{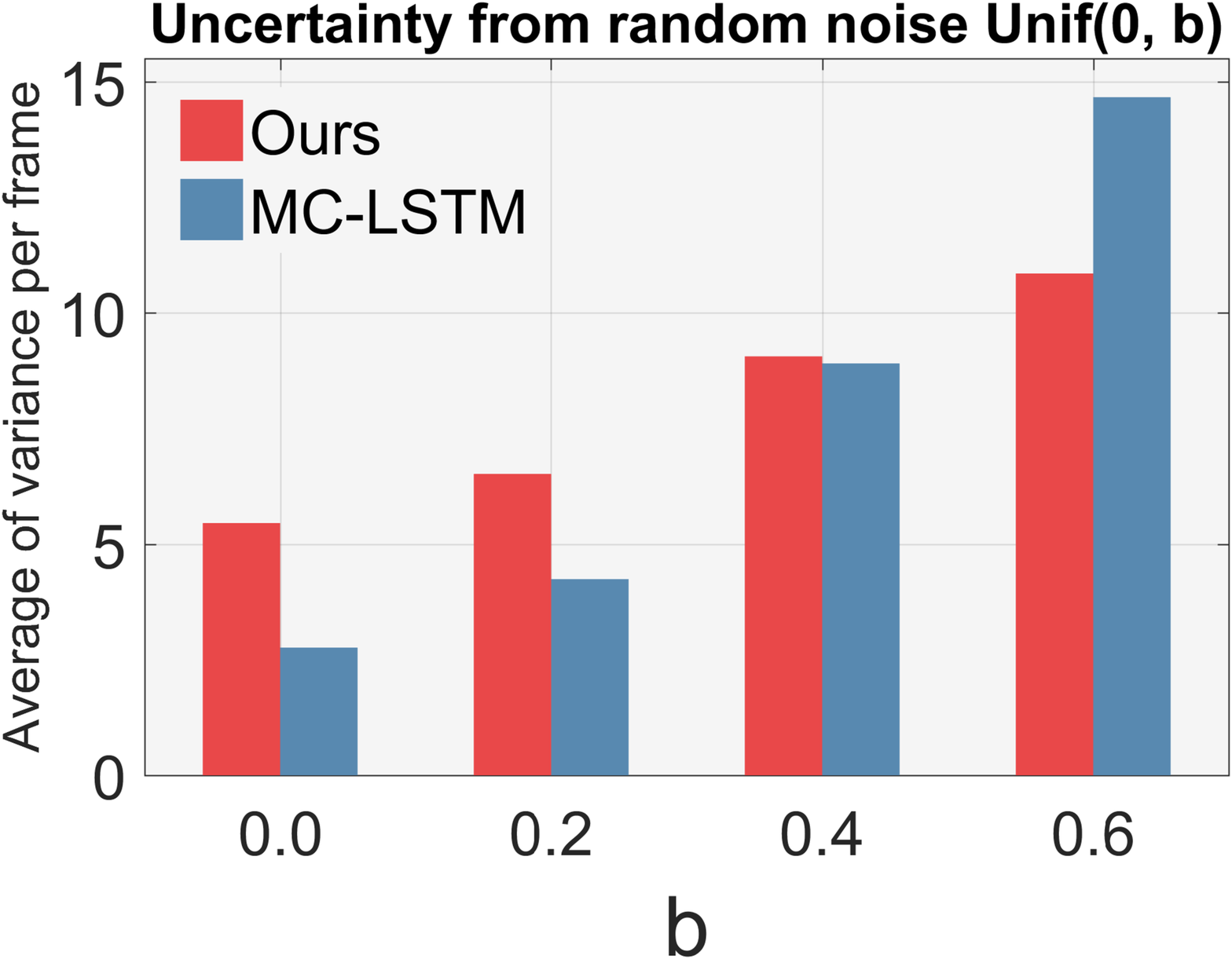}
				\end{minipage}
	\end{minipage}
	\caption{\label{fig:deviation}Predictions and uncertainties (frames 11, 15, and 20) from testing varying deviations from trained trajectories (first of four rows, blue). %using a model trained only on $\theta=20$ (blue line of Fig.~\ref{fig:traj}).
	Top: angle. Middle: speed. Bottom: pixel-level noise. Right: the average sum of per frame pixel-level variance using \GRUNAME{} and MC-LSTM.}
\end{figure*}

\noindent\textit{Results.} For varying angle and speed deviations, (top/middle of Fig.~\ref{fig:deviation}), the predictions look visually sensible, but they \textit{do not} actually follow the ground truth trajectories.
%We notice that despite how 'good looking' the predictions are under various angle and speed changes (Top and Bottom of Fig.~\ref{fig:deviation}), the predictions \textit{do not} actually follow the ground truth trajectories.
%This relates to our previous question: How certain are we about our visually plausible results?
%Without manually analyzing the predictions,
We can quantify this directly by the sum of pixel-level variance per frame (averaged over time and testset) as shown in the right of Fig.~\ref{fig:deviation}. The \textit{uncertainty increases as the angle/speed deviation increases} for both \GRUNAME{} and MC-LSTM. The level of pixel-noise (bottom of Fig.~\ref{fig:deviation}) also increases model uncertainty (note the trajectories remain the same). These observations exactly demonstrate the usefulness of uncertainty when the prediction output itself is visually sensible.

%(especially when visually sensible) does not provide the correctness.
%\begin{wrapfigure}[11]{l}{0.3\textwidth}%[!t]
%	\centering
%	\vspace{-20pt}
%	\includegraphics[width=0.3\columnwidth]{mc/angle_test1_3_ours}\\ \vspace{5pt}
%	\includegraphics[width=0.3\columnwidth]{mc/angle_test1_3_mc}
%	\caption{\footnotesize\label{fig:mc_compare} Comparison of uncertainty of \GRUNAME{} (top) and MC-LSTM (bottom).}
%\end{wrapfigure}
From a practical perspective, the uncertainty inference should not sacrifice computational speed, e.g., realtime safety of an autonomous vehicle.
%For instance, acknowledging the depth uncertainty of an autonomous vehicle's vision system absolutely must be done close to real-time for maximal safety.
%Thus, inference time is a crucial practical aspect of uncertainty system, and
With respect to this crucial aspect, \GRUNAME{} greatly benefits from its \textit{sampling-free} procedure: each epoch ($30$ sequences) takes $\sim$3 seconds while MC-LSTM with a Monte Carlo rate of $50$ requires $\sim$40 seconds ($> 10$ times \GRUNAME{}) despite their comparable qualitative performance.
The MC sampling rate for these methods \textit{cannot} simply be decreased: uncertainty will be underestimated. Further, theoretical analysis may be necessary to determine the rate of convergence to the true model uncertainty, if convergence is guaranteed at all. With \GRUNAME{}, we compute this model uncertainty \textit{in closed form}, without the need for any heavy lifting from large sample analysis.
% he sampling size of MC-LSTM (or any other sampling-based schemes) and underestimate uncertainty, especially in more complex settings where heavy theoretical analysis may be necessary from large samples to determine the rate of convergence to the true model uncertainty with guaranteed convergence.
%With EF-GRU, we compute this model uncertainty \textit{in closed form}, without the need for any heavy lifting from large sample analysis compared to MC-LSTM which only \textit{approximates} the model uncertainty for a given sample

\noindent\textit{Qualitative Comparisons.} We demonstrated that the uncertainty measures of both SP-GRU and MC-LSTM behave similarly such that the increasing trajectory deviations had also increased the uncertainties (Fig.~\ref{fig:deviation}). However, the absolute magnitudes of these measures do not necessarily translate directly to the quality of uncertainty. For instance, given the same unfamiliar input, if the uncertainty of SP-GRU is higher than that of MC-LSTM, that does not necessarily mean that the uncertainty estimate of SP-GRU is better than that of MC-LSTM. Instead, the relative measure of uncertainty within each model is of more interest which demonstrates how each model appropriately detects the degree of uncertainty given various inputs.

\begin{figure}[h]
	\centering
	\includegraphics[width=0.3\textwidth]{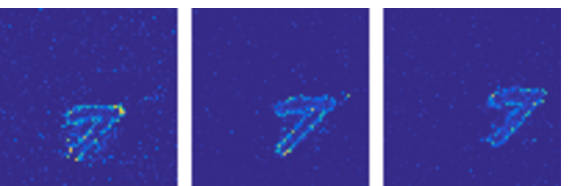} \hspace{20pt}
	\includegraphics[width=0.3\textwidth]{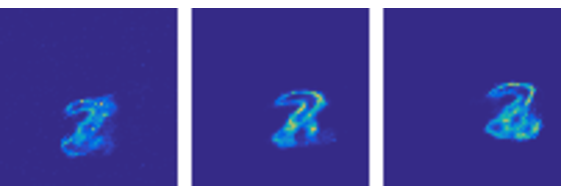}
	\caption{\label{fig:map} Uncertainty maps of SP-GRU (left) and MC-LSTM (right) of three identical frames given a trajectory with deviated angle.}
\end{figure}

Thus, in Fig.~\ref{fig:map}, we qualitatively show that the uncertainty measures of the models, despite the small gaps between their quantities, agree with each other. In other words, we see how the uncertainty maps from SP-GRU and MC-LSTM look similar to each other, implying that (1) these models interpret the uncertainty similarly (which is reasonable given the similarity between their basic cell structures) and (2) the between-model uncertainty gaps are of less significance.

\noindent\textbf{Random Moving MNIST.} We perform two learning tasks on training sequences of $20$ frames of two moving moving digits. Given the first 10 input frames, (1) reconstruct the same input frames and (2) predict the next 10 frames. We use the same test set provided by \cite{srivastava2015unsupervised} to facilitate model comparison.
%For these tasks, the \GRUNAME{} is trained separately with a Gaussian exponential family distribution determining the natural parameters.
%\vspace{-10pt}

%\begin{figure*}[!t]
%\centering
%$\overbrace{\hspace{0.495\textwidth}}^{\text{Input and Decoder Estimate}}$ $\overbrace{\hspace{0.495\textwidth}}^{\text{Prediction Estimate}}$ \\
%\includegraphics[width=\textwidth]{1digits/1977_gt} \\ \smallskip
%\includegraphics[width=\textwidth]{1digits/1977_mean} \\ \smallskip
%\includegraphics[width=\textwidth]{1digits/1977_var} \\ \bigskip
%\includegraphics[width=\textwidth]{3digits/127_gt} \\ \smallskip
%\includegraphics[width=\textwidth]{3digits/127_mean} \\ \smallskip
%\includegraphics[width=\textwidth]{3digits/127_var}
%\caption{\footnotesize\label{fig:mnist2} NP-GRU decoder + predictor results on out of domain samples.}
%\end{figure*}
%\input{fig_speed_var.tex}
%\input{fig_noise_var.tex}

\begin{figure*}[!t]
	\centering
	$\overbrace{\hspace{0.495\textwidth}}^{\text{Input and Decoder Estimate}}$ $\overbrace{\hspace{0.495\textwidth}}^{\text{Prediction Estimate}}$ \\
	\includegraphics[width=\textwidth]{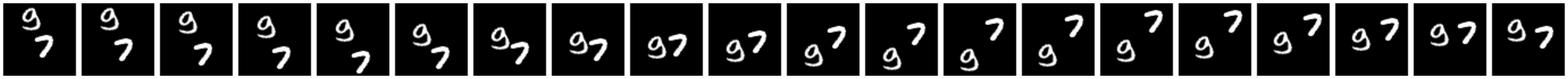} \\ \vspace{1pt}
	\includegraphics[width=\textwidth]{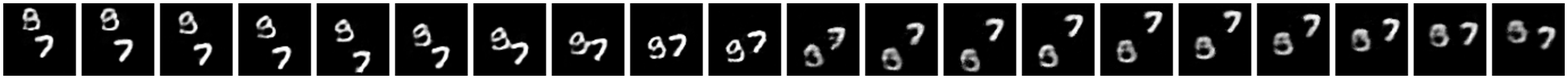} \\ \vspace{1pt}
	\includegraphics[width=\textwidth]{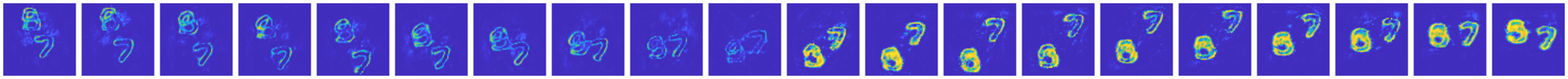} \\ \vspace{2pt}
	\includegraphics[width=\textwidth]{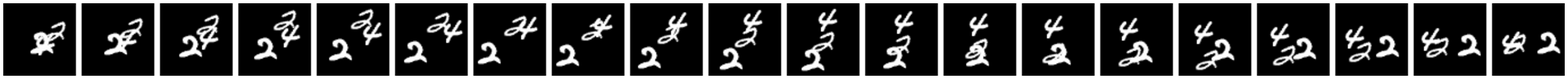} \\ \vspace{1pt}
	\includegraphics[width=\textwidth]{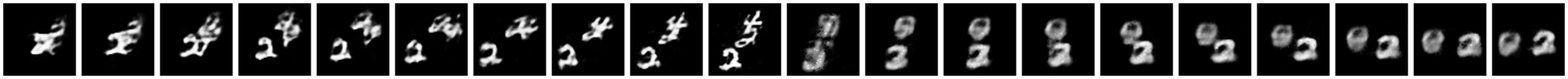} \\ \vspace{1pt}
	\includegraphics[width=\textwidth]{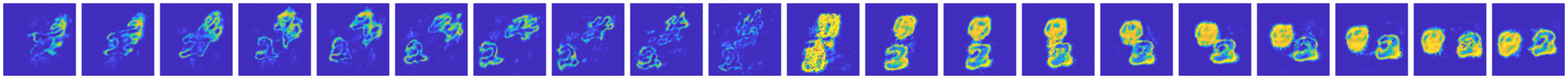}
	\caption{\label{fig:mnist1}\GRUNAME{} decoder and predictor results. Each row corresponds to the ground truth, mean prediction, and variance estimate respectively.}
		  		\vspace{-10pt}
\end{figure*}

\noindent\textit{Results.} A test sample is given in Fig.~\ref{fig:mnist1} (top three rows). As expected, the quality of the input reconstruction diminishes as the network progresses further into the past. We see a reflected, but similar behavior with the predictor model. Notice that these qualitative interpretations are \textit{directly} quantifiable through the variance output.%; the third line of each sample shows this progression for a specific sample, and the seventh row shows how this uncertainty behaves averaged over the entire test set. In general, the model is confident in empty regions \textit{and in digit interiors} (low variance), while most uncertain around digit edges and blurry reconstructions (high variance).
We briefly evaluate how well \GRUNAME{} is able to perform on \textit{out-of-domain} samples (Fig. \ref{fig:mnist1}, bottom three rows). Models deployed in real-world settings may not realistically be able to determine if a sample is far from their training distributions. 
\begin{wraptable}[12]{r}{0.5\textwidth}%[!t]
	\centering
\caption{\label{fig:table} Average cross entropy test loss per image per frame on Moving MNIST.}
	\begin{tabular}{lr}%c}
		%		& Test & \\
		Model & Test Loss \\%& $\sim$ \\%Uncertainty\\
		\toprule\toprule
		\cite{srivastava2015unsupervised} Srivastava et al. & 341.2\\%& $\times$ \\
		\cite{xingjian2015convolutional} Xingjian et al. & 367.1\\%& $\times$ \\
		\cite{de2016dynamic} Brabandere et al. & 285.2\\%& $\times$ \\
		\cite{ghosh2016contextual} Ghosh et al. & 241.8\\%& $\times$ \\
		%		Cricri et al. \cite{cricri2016video} & 187.7& N \\
		%		Kalchbrenner et al. \cite{kalchbrenner2016video} & 87.6& N \\
		\GRUNAME{} & 277.1 \\%& \textbf{\checkmark} \\
		\bottomrule\bottomrule
	\end{tabular}
\end{wraptable}
However, with our specific modeling of uncertainty, we would expect that images or sequences distant from the training data will exhibit high variance.
 We construct sequences of 3 moving digits. In this case, future reconstruction is generally quite poor. As has been observed in previous work \cite{srivastava2015unsupervised}, the model attempts to hallucinate two digits. Our model is \textit{aware of this issue}: the variance for a large number of pixels is extremely high, \textit{regardless of if the digits overlap}. Even in ``easy" cases where there is little overlap and digits are relatively clear, the model has some sense of the particular input sample and recovery being far from its learning distribution.

We compare our method to previous work in Table \ref{fig:table}. 
%Here, we only compute the cross entropy with the computed mean for each image frame. 
\GRUNAME{} with a basic predictor network setup (Fig.~\ref{fig:table}) performs comparably or better than other methods that do \textit{not} provide model uncertainty. In these works, model performance often benefits from their specific network structure: encoder-predictor composite models \cite{srivastava2015unsupervised}, generative adversarial networks \cite{ghosh2016contextual}, and external weight filters \cite{de2016dynamic}. Further, more recent models \cite{cricri2016video,kalchbrenner2016video} have achieved better results with large, more sophisticated pipelines. Extending \GRUNAME{} to such setups becomes a reasonable modification, providing model uncertainty \textit{without} sacrificing performance.

\begin{figure}[!h]
	\includegraphics[width=\textwidth]{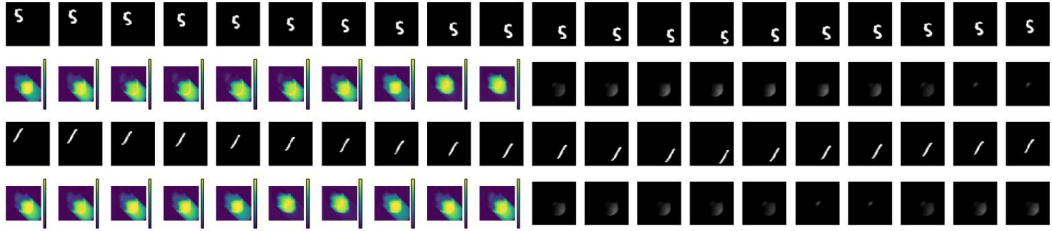}
	\caption{\label{fig:dmm} Results for a single-digit fixed trajectory after 646K iterations using the DMM model. Top: Ground Truth Trajectory. Bottom Right: Predicted trajectory of last 10 frames given first 10. Bottom Left: Uncertainty estimation of predicted trajectory.}
\end{figure}

\noindent\textbf{Other Methods of Measuring Uncertainty.} Deep Markov Models \cite{krishnan2017structured} introduced recently naturally give rise to a probabilistic interpretation of predictions from deep temporal models. However, upon application of this model to Moving MNIST we were unable to obtain any reasonable prediction, across a range of hidden dimension sizes and trajectory complexities, even with significant training time (days vs. hours for \GRUNAME{}). Shown in Fig.~\ref{fig:dmm} are results using a hidden dimension size of 1024, with 100 dimensional hidden state. We note that the experimental setups described in \cite{krishnan2017structured} are small in dimension and complexity compared to Moving MNIST, and it may be the case that small technical development with DMMs may lead to promising and comparable uncertainty results.

%While many of the methods compared here train both the decoder and predictor concurrently, we observe that the our disjoint NP-GRU competes favorably even without the additional information provided by a coupled setup.

%\subsection{Other Methods of Modeling Uncertainty}
%-description of MC dropout, and then...

%At test time, 3 secs vs 30 secs...
%Critically, methods such as MCD are only \textit{approximating} the model uncertainty for a given sample. In these cases heavy theoretical analysis may be necessary to determine the rate of convergence to %the true model uncertainty, if convergence is guaranteed at all. With NPGRU, we compute this model uncertainty \textit{in closed form}, without the need for any heavy lifting from large sample analysis.

% THIS WENT TO SUPPLEMENT
%We note that Deep Markov Models \cite{krishnan2017structured} introduced recently natural give rise to a probabilistic interpretation of predictions from deep temporal models, but upon application of this %model to Moving MNIST we were unable to obtain any reasonable prediction, even with large dimensional latent variables over significant training time. See supplement for details.

\vspace{-5pt}
\section{Conclusion}
\vspace{-5pt}
Recent developments in vision and machine learning suggest that in the near term future,
network-based algorithms will play an increasing role in systems deployed in healthcare, economics, scientific studies (outside of computer science) and even policy making. 
But to facilitate this progression, one of potentially many requirements will be the ability 
of the models to {\em meaningfully reason about uncertainty}. Complementary to 
the developing body of work on Bayesian perspectives on deep learning, this paper shows 
how a mix of old and new ideas can enable deriving uncertainty estimates for a powerful class of models, 
GRUs, which is easily extensible to other sequential models as well. While the low-level 
details of the approach are involved, it is intuitive and works surprisingly well in practice. 
We first show applications on a standard dataset used in sequential models 
where our results are competitive with other algorithms, while offering uncertainty as a natural byproduct.
%Later, we demonstrated an interesting application of these ideas 
%to the analysis of brain imaging data which yields promising initial results.
The 
TensorFlow implementation will be publicly available.

%\newpage
\bibliographystyle{splncs}
\bibliography{eccv2018_uncertainty}

\newpage
\begin{appendices}
\section{Nonlinear Transformations of Other Exponential Family Distributions}

We show the activation functions (nonlinear transformations) of some common exponential family distributions. As we mentioned in the main text, the closed form solutions of the nonlinear transformations (Sec.~3.3) of several exponential family distributions involves a monotonically increasing and bounded 'activation-like' mapping $f(x) = a - b \exp(-\gamma d(x))$ where $d(x)$ is an arbitrary activation of choice with appropriate constants $a$, $b$ and $\gamma$. Using $f(x)$, the goal is to perform nonlinear transformations on the pre-activation output $o$ with respect to mean $m$ and variance $s$, equivalent to the following marginals:
\begin{equation}
a_m = \int f(o)p_O(o \ | \ o_\alpha, o_\beta) do \quad \textrm{and} \quad  a_s = \int f(o)^2 p_O(o \ | \ o_\alpha, o_\beta) do - a_m^2
\end{equation}
which we express in closed forms. We refer to \cite{wang2016natural} to show the derivations for Gamma, Poisson and Gaussian distributions.

\subsection{Gamma Distribution}
The probability density function (pdf) of the Gamma distribution given parameters $\alpha>0$ and $\beta>0$ (not to be confused with natural parameters as these are conventional notations) is
\begin{equation*}
p_X(x \ | \ \alpha, \beta) = \frac{\beta^\alpha}{\Gamma(\alpha)} x^{\alpha - 1}\exp(-\beta x)
\end{equation*}
for the support $x \in (0,\infty)$ and $\Gamma(\alpha) = \int_0^\infty x^{\alpha - 1} \exp(-x)dx$. Then, let $c = a = b$, $c > 0$ and $d(x) = x$ to get $f(x) = c(1-\exp(-\gamma x))$. We first perform a nonlinear transformation with respect to $m$ as follows:
\begin{align*}
a_m &= \int f(o)p_O(o \ | \ o_\alpha, o_\beta) do \\
&= \int_{o=0}^\infty c(1-\exp(-\gamma o)) \frac{{o_\beta}^{o_\alpha}}{\Gamma(o_\alpha)} o^{o_\alpha - 1}\exp(-o_\beta \odot o) do \\
&= c\int_{o=0}^\infty \frac{{o_\beta}^{o_\alpha}}{\Gamma(o_\alpha)} o^{o_\alpha - 1}\exp(-o_\beta \odot o) do \\
&\qquad - c\int_0^\infty \frac{{o_\beta}^{o_\alpha}}{\Gamma(o_\alpha)} o^{o_\alpha - 1}\exp(-(\gamma o+o_\beta) \odot o) do\\
&= c\left[1 - \frac{{o_\beta}^{o_\alpha}}{\Gamma(o_\alpha)}\int_0^\infty  o^{o_\alpha - 1}\exp(-(\gamma o+o_\beta) \odot o) do\right]\\
&= c\left[1 - \frac{{o_\beta}^{o_\alpha}}{\Gamma(o_\alpha)}\odot \Gamma(o_\alpha) \odot (o_\beta + \gamma) ^{-o_\alpha} \right]\\
&= c\left[1 - \frac{{o_\beta}^{o_\alpha}}{(o_\beta + \gamma) ^{o_\alpha}} \right]\\
&= c\left[1 - \left( \frac{{o_\beta}}{o_\beta + \gamma} \right)^{o_\alpha} \right]
\end{align*}
and for the variance,
\begin{align*}
a_s &= \int f(o)^2p_O(o \ | \ o_\alpha, o_\beta) do - a_m^2\\
&= \left[  \int_{o=0}^\infty c^2(1-2\exp(-\gamma o)+\exp(-2\gamma o)) \frac{{o_\beta}^{o_\alpha}}{\Gamma(o_\alpha)} o^{o_\alpha - 1}\exp(-o_\beta \odot o) do \right] - a_m^2\\
&= c^2\left[ 1 - 2\frac{{o_\beta}^{o_\alpha}}{\Gamma(o_\alpha)} \odot \Gamma(o_\alpha)\odot(o_\beta + \gamma)^{-o_\alpha} + \frac{{o_\beta}^{o_\alpha}}{\Gamma(o_\alpha)}\odot \Gamma(o_\alpha) \odot (o_\beta + 2\gamma) ^{-o_\alpha} \right] \\
&\qquad - a_m^2\\
&= c^2\left[ 1 - 2\frac{{o_\beta}^{o_\alpha}}{(o_\beta + \gamma) ^{o_\alpha}} + \frac{{o_\beta}^{o_\alpha}}{(o_\beta + 2\gamma) ^{o_\alpha}} \right] - a_m^2\\
&= c^2\left[\left( \frac{{o_\beta}}{o_\beta + 2\gamma} \right)^{o_\alpha} - \left( \frac{{o_\beta}}{o_\beta + \gamma} \right)^{2o_\alpha} \right]
\end{align*}
for $c > 0$ and $\gamma > 0$ where $c=1$ and $\gamma=1$ are generally good choices that resemble $\tanh$.

\subsection{Poisson Distribution}
The pdf of the Poisson distribution over the support $x \in \{0,1,2,\dots\}$ with a parameter $\lambda > 0$ is
\begin{equation}
p_X(x \ | \ \lambda) = \frac{\lambda^x \exp(-\lambda)}{x!}.
\end{equation}
Then, let $c = a = b$, $c > 0$ and $d(x) = x$ to get $f(x) = c(1-\exp(-\gamma x))$. The nonlinear transformation on $o$ to obtain $a_m$ is as follows:
\begin{align*}
a_m &= \sum_{x=0}^\infty f(x)p_O(o \ | \ o_\alpha, o_\beta) \\
&= \sum_{x=0}^\infty c(1-\exp(-\gamma x)) \frac{o_\alpha^x \exp(-o_\alpha)}{x!} \\
&= c\sum_{x=0}^\infty \frac{o_\alpha^x \exp(-o_\alpha)}{x!} - c\sum_0^\infty \exp(-\gamma x)\frac{o_\alpha^x \exp(-o_\alpha)}{x!} \\
&= c - c(\exp(-o_\alpha))\sum_{x=0}^\infty \frac{o_\alpha^x \exp(-\gamma x)}{x!} \\
&= c[1 - \exp(-o_\alpha) \exp ( \exp(-\gamma)o_\alpha))]
\end{align*}
and for the variance,
\begin{align*}
a_s &= \sum_{x=0}^\infty f(x)p_O(o \ | \ o_\alpha, o_\beta) - a_m^2\\
&= \left[ \sum_{x=0}^\infty c^2(1-2\exp(-\gamma x)+\exp(-2\gamma x)) \frac{o_\alpha^x \exp(-o_\alpha)}{x!}  \right] - a_m^2 \\
&= c^2 \sum_{x=0}^\infty \frac{o_\alpha^x \exp(-o_\alpha)}{x!} - 2c^2\exp(-o_\alpha)\sum_{x=0}^\infty \frac{o_\alpha^x \exp(-\gamma x)}{x!}\\
&\qquad + c^2\exp(-o_\alpha)\sum_{x=0}^\infty \frac{o_\alpha^x \exp(-2\gamma x)}{x!} - a_m^2 \\
&= -c^2\exp(2(\exp(-\gamma)-1)o_\alpha) + c^2 \exp((\exp(-2\gamma)-1)o_\alpha)\\
&= c^2[\exp((\exp(-2\gamma)-1)o_\alpha)  -  \exp(2(\exp(-\gamma)-1)o_\alpha)] 
\end{align*}
for $c > 0$ and $\gamma > 0$.

\subsection{Gaussian Distribution}
Here, we provide details of the nonlinear transformation on the Gaussian distribution (Eq.~(5,6,7,8) of main text). Note that our goal is to compute
\begin{equation*}\label{eq:gaussian_sig}
a = \int_{-\infty}^{\infty} f(x) N(x \ | \ m, s^2)dx
\end{equation*}
for some nonlinear function $f(x)$, mean $m$ and variance $s^2$. First, we consider $f(x) = \sigma(x)$. Then, Eq.~\ref{eq:gaussian_sig} is the logistic-normal integral:
\begin{equation*}
a = \int_{-\infty}^{\infty} \sigma(x) N(x \ | \ m, s^2)dx = \int_{-\infty}^{\infty} \frac{1}{1+e^{-x}} \frac{1}{\sqrt{2\pi s^2}} \exp\left(-\frac{(x-m)^2}{2s^2}\right)dx
\end{equation*}
which does not have a closed form solution. Now, we use the fact that a probit function
\begin{equation*}
\Phi(x) = \int_{-\infty}^{x} N(z \ | \ 0,1) dt
\end{equation*}
can be used to approximate a sigmoid function such that
\begin{equation*}
\sigma(x) \approx \Phi(\zeta x)
\end{equation*}
for $\zeta^2 = \pi/8$. Further, we know that 
\begin{equation*}
\int_{-\infty}^{\infty} \Phi(x) N(x \ | \ m, s^2)dx = \Phi \left( \frac{m}{\sqrt{1+s^2}} \right)
\end{equation*}
so the nonlinear transformation on $o$ with respect to $m$ is
\begin{equation*}
a_m = \int \sigma(o) N(o \ | \ o_\alpha, diag(o_\beta)) do \approx \Phi \left( \frac{o_\alpha}{\sqrt{\zeta^{-2} + o_\beta}} \right) = \sigma \left( \frac{o_\alpha}{\sqrt{1 + \zeta^2 o_\beta}} \right)
\end{equation*}
for $\zeta^2 = \pi/8$. Similarly, for the variance, since
\begin{equation*}
\sigma(x)^2 = \Phi(\zeta\nu(x + \omega))
\end{equation*}
for $\nu = 4-2\sqrt{2}$ and $\omega = -\log(\sqrt{2}+1)/2$, we see that
\begin{align*}
a_s &= \int \sigma(o)^2 N(o \ | \ o_\alpha, diag(o_\beta)) do - a_m^2 \\
&= \Phi \left( \frac{\nu(o_\alpha + \omega)}{\sqrt{\zeta^{-2} + \nu^2o_\alpha} } \right) - a_m^2 \\
&= \sigma \left( \frac{\nu(o_\alpha + \omega)}{\sqrt{1 + \zeta^2\nu^2o_\alpha} } \right) - a_m^2
\end{align*}
for $\zeta^2 = \pi/8$.
\\ \\
The hyperbolic tangent function can be derived in a similar way since $\tanh(x) = 2\sigma(2x)-1$. Thus, for $f(x) = \tanh(x)$ over the support $x \in (-\infty,\infty)$,
\begin{align*}
a_m &= \int \tanh(o) N(o \ | \ o_\alpha, diag(o_\beta))do \\
&= \int (2\sigma(2o)-1) N(o \ | \ o_\alpha, diag(o_\beta))do \\
&= 2\int \sigma(2o) N(o \ | \ o_\alpha, diag(o_\beta))do - \int_{-\infty}^{\infty} \sigma(2o) N(o \ | \ o_\alpha, diag(o_\beta))do\\
&= 2\int \sigma(2o) N(o \ | \ o_\alpha, diag(o_\beta))do - 1\\
&\approx 2\int \Phi(2o) N(o \ | \ o_\alpha, diag(o_\beta))do - 1\\
&= 2 \Phi \left( \frac{2 \zeta o_\alpha}{\sqrt{1 + 4 \zeta^2 o_\beta}} \right) - 1 \\
&\approx 2 \sigma \left( \frac{2 o_\alpha}{\sqrt{1 + 4 \zeta^2 o_\beta}} \right) - 1 \\
&= 2 \sigma \left(\frac{o_\alpha}{\sqrt{\frac{1}{4} + \zeta^2 o_\beta}}\right) - 1
\end{align*}
and for the variance,
\begin{align*}
a_s &= \int \tanh(o)^2 N(o \ | \ o_\alpha, diag(o_\beta))do - a_m^2 \\
&= \int (4\sigma(2o)^2 - 4 \sigma(2o) + 1) N(o \ | \ o_\alpha, diag(o_\beta))do - a_m^2 \\
&\approx \int (4\Phi(\zeta \nu(o + \omega)) - 4 \sigma(2o) + 1) N(o \ | \ o_\alpha, diag(o_\beta))do - a_m^2 \\
&= \int 4\Phi(\zeta \nu(o + \omega))N(o \ | \ o_\alpha, diag(o_\beta))do - \int 4 \sigma(2o) N(o \ | \ o_\alpha, diag(o_\beta))do \\
&\qquad + 1 - a_m^2 \\
&= 4 \Phi \left( \frac{\nu(o_\alpha + \omega)}{\sqrt{\zeta^{-2} + \nu^2 o_\beta}} \right) - 2 \sigma \left(\frac{o_\alpha}{\sqrt{\frac{1}{4} + \zeta^2 o_\beta}}\right) - 3 - a_m^2 \\
&\approx 4 \sigma \left( \frac{\nu(o_\alpha + \omega)}{\sqrt{1 + \zeta^2\nu^2 o_\beta}} \right) - a_m^2 - 2 a_m - 1 
\end{align*}
where $\nu = 2(4-2\sqrt{2})$ and $\omega = -\log(\sqrt{2}+1)/2$.
\end{appendices}

\end{document}